%% file: main.tex
\newcommand{\modelname}{\textsc{DualVLA}}
\newcommand{\decay}{Action Degeneration}
\newcommand{\evalname}{\textit{VLA Score}}
\documentclass[10pt,twocolumn,letterpaper]{article}
\usepackage{paralist}
\usepackage{multirow} 
\usepackage{listings}
\usepackage{marvosym}  

\usepackage[most]{tcolorbox} 
\usepackage[pagenumbers]{cvpr} 

\input{preamble}
\definecolor{cvprblue}{rgb}{0.21,0.49,0.74}
\usepackage[pagebackref,breaklinks,colorlinks,allcolors=cvprblue]{hyperref}
\usepackage{algorithm}
\usepackage{algorithmic}
\usepackage[dvipsnames, svgnames, x11names]{xcolor}
\usepackage{xcolor}  
\usepackage{colortbl}  
\usepackage{natbib}
\usepackage{float}
\setcitestyle{square, comma, numbers,sort&compress, super}
\def\paperID{3872} 
\def\confName{CVPR}
\def\confYear{2026}

\title{DualVLA: Building a Generalizable Embodied Agent \\via Partial Decoupling of Reasoning and Action}

\begin{document}
\author{
\textbf{Zhen Fang$^{1*}$} \quad
\textbf{Zhuoyang Liu$^{2*}$} \quad
\textbf{Jiaming Liu$^{2\dagger}$} \quad
\textbf{Hao Chen$^{3}$} \quad
\textbf{Yu Zeng$^{1}$} \\
\textbf{Shiting Huang$^{1}$} \quad
\textbf{Zehui Chen$^{1\dagger}$} \quad
\textbf{Lin Chen$^{1}$} \quad
\textbf{Shanghang Zhang$^{2\,\textrm{\Letter}}$} \quad
\textbf{Feng Zhao$^{1\,\textrm{\Letter}}$}\\
$^{1}$MoE Key Laboratory of Brain-inspired Intelligent Perception and Cognition,\\University of Science and Technology of China \quad
$^{2}$ State Key Laboratory \\of Multimedia Information Processing, School of Computer Science, \\Peking University \quad $^{3}$ CUHK\\
{\small $^{*}$ Equal Contribution. \quad $^{\dagger}$ Project Lead. \quad $^{\textrm{\Letter}}$ Corresponding Authors.}
}

\twocolumn[{%
\renewcommand\twocolumn[1][]{#1}%
\maketitle

\centering
\includegraphics[width=\linewidth]{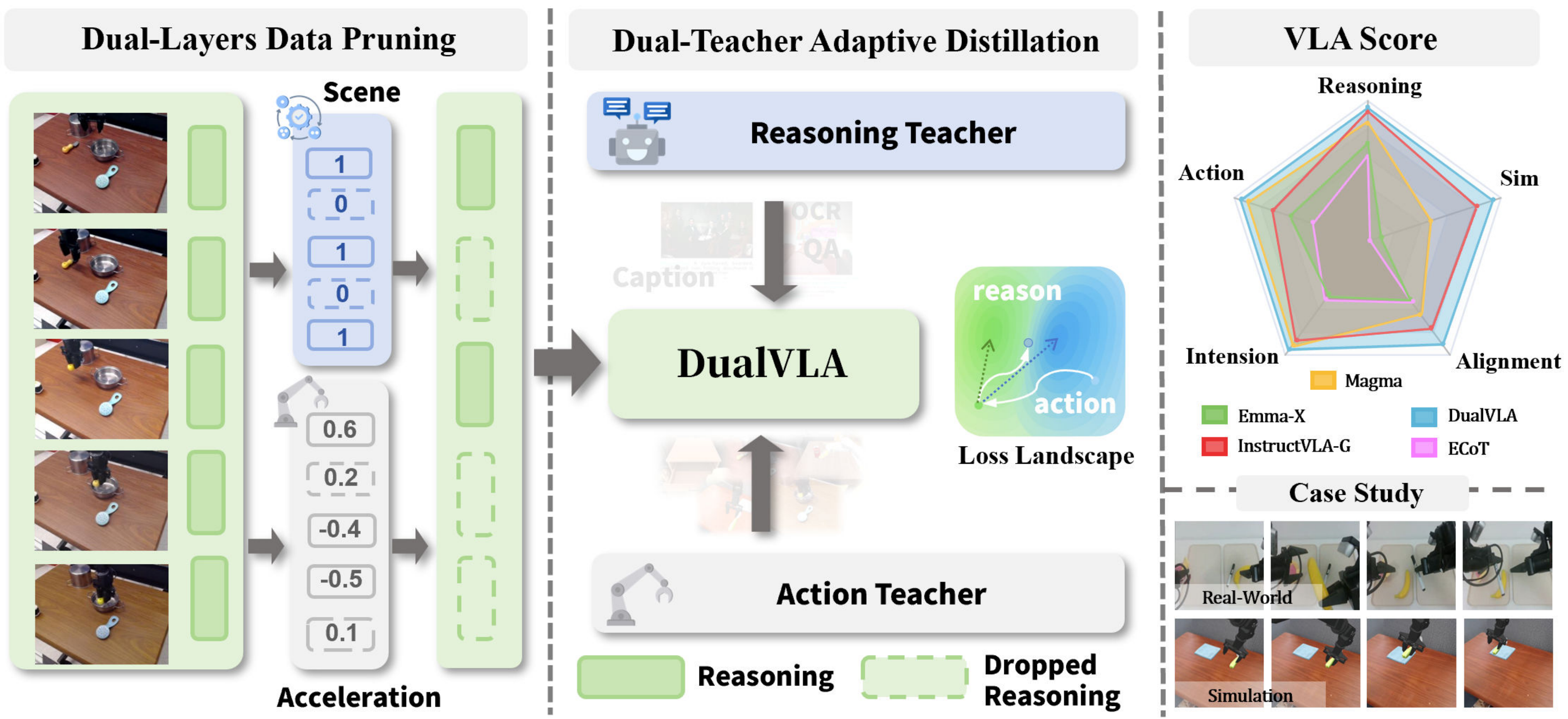}

\captionof{figure}{{\modelname} first constructs a sparse, information-dense embodied reasoning dataset by combining video event prediction with kinematic cues, mitigating the negative impact of redundant reasoning on action generation. It then adopts a dual-teacher strategy: an action teacher offering fine-grained supervision for manipulation, and a reasoning teacher maintaining general reasoning capability. Together, these components enable {\modelname} to achieve strong performance in both simulation and real-world robotic evaluations.\vspace{1em}}
\vspace{+0.3cm}
\label{fig:teaser}
}
]

\input{sec/0_abstract}    
\input{sec/1_introv2}

\input{sec/2_related_work}
\input{sec/4_methd}
\input{sec/5_experiments}
\input{sec/6_conclusion}

{
    \small
    \bibliographystyle{ieeenat_fullname}
    \bibliography{main}
}
\input{sec/99_supp}

\end{document}

%% file: sec/0_abstract.tex
\begin{abstract}
To build a generalizable Vision–Language–Action (VLA) model with reasoning capability, a common approach is to first train a specialist VLA on robot demonstrations to acquire reliable manipulation skills, and then introduce mixed annotated-robot data with multimodal data to restore general reasoning.
However, we observe that the resulting reasoning VLA exhibits degraded action performance compared to the specialist VLA before fine-tuning.
We define this phenomenon as \textbf{action degeneration}.
To tackle this issue, we propose {\modelname}, which improves action performance through carefully designed post-training while preserving the reasoning ability.
We first propose a dual-layer data pruning method to remove redundant embodied reasoning and alleviate its adverse guidance on action learning.
To further enhance the model's action generation capabilities, we adopt a dual-teacher adaptive distillation strategy that assigns different supervision signals to different data domains and maintains its reasoning ability. 
To fill the evaluation gap of generalist VLAs, we introduce VLA Score, which decouples VLA capabilities into reasoning, intention, action, and alignment, enabling a more fine-grained evaluation.
Experiments show that {\modelname} achieves an average success rate of 61.0 in SimplerEnv and an average score of 65.4 across eight competitive multimodal benchmarks, demonstrating a stronger balance between action execution and multimodal understanding.
\href{https://costaliya.github.io/DualVLA/}{\textcolor{blue}{Project Website}}.
\end{abstract}

%% file: sec/1_introv2.tex
\section{Introduction}
\label{sec:intro}
%
An embodied agent is an AI system that perceives, reasons, and acts within complex environments, grounding intelligence in real-world interaction. Vision–Language–Action(VLA) models represent a key step toward such agents by leveraging the rich priors of Vision–Language Models (VLMs) and large-scale robotic datasets~\cite{oxe-dataset,bridgev2,calvin} to map observations and instructions to control signals. Recent advances~\cite{pi0,openvla,cogact,hybridvla} show strong manipulation accuracy, generalization, and long-horizon capability, highlighting the promise of VLAs for embodied intelligence.

Previous specialist VLAs~\cite{openvla,oxe-dataset, spatialvla, tracevla} fine-tune a VLM on robotic datasets and thus achieve strong manipulation performance, though their multimodal reasoning ability remains limited. To enrich the reasoning capacity of VLAs, which helps models leverage VLM priors during manipulation and interpret complex scenes. Recent works augment robot trajectories with reasoning annotations and mix them with multimodal corpora during finetuning~\cite{ecot,emma-x,ecot-lite,fast-ecot,instructvla,chatvla}. This paradigm provides a practical route toward building reasoning VLAs.
Building on this direction, we observe that enhancing a specialist VLA with additional reasoning may reduce its manipulation performance, a phenomenon we refer to as action degeneration. This suggests that reasoning and action rely on shared internal representations, and that reasoning-oriented supervision can inadvertently reshape the model's visuomotor behavior. Importantly, this decline arises even when the training data becomes larger and more diverse. Such a deviation from the expected trend predicted by the scaling law~\cite{scalinglaw} highlights a fundamental challenge: increasing data volume alone is insufficient unless the supervision signals for reasoning and action are properly balanced.

To address this issue, we propose {\modelname}, a reasoning VLA equipped with a dedicated post-training framework that enhances action performance while preserving multimodal reasoning capabilities. Our central insight is that action degradation arises from two factors: the misleading influence of repetitive, low-entropy embodied reasoning in the training data, and the lack of differentiated, fine-grained supervision for simultaneously learning reasoning and action.
Guided by this insight, we first observe that redundant embodied reasoning can bias the optimization process and deteriorate action execution. To mitigate this effect, we introduce a dual-layer data pruning strategy that leverages both embodiment cues and scene-level event changes to remove unnecessary reasoning segments while retaining action-critical content.
To explicitly strengthen manipulation ability, we further utilize the specialist VLA as a natural provider of high-quality action supervision. Building on mix training, we develop a dual-teacher adaptive distillation strategy that assigns distinct soft-label supervision to robot data and multimodal reasoning data, enabling the model to learn both capabilities under balanced and fine-grained guidance.
Finally, we note that existing VLAs evaluations do not disentangle reasoning from action and therefore cannot faithfully reflect the performance trade-offs observed in reasoning VLAs. To address this limitation, we introduce {\evalname}, the first evaluation pipeline tailored for reasoning-capable VLAs. By incorporating a strong VLM as an evaluator, {\evalname} provides comprehensive assessment along four dimensions—action, reasoning, intention, and reasoning–action alignment—and brings the MLLM-as-a-Judge paradigm into the evaluation of VLA systems.

In summary, our contributions are three-fold:
\begin{itemize}
    \item     
We propose {\modelname}, a reasoning VLA with a post-training framework that decouples reasoning and action learning at the data and loss levels. Through dual-layer reasoning pruning and dual-teacher adaptive distillation, {\modelname} effectively preserves multimodal reasoning while improving manipulation ability.
    \item    
We introduce {\evalname}, the first evaluation framework tailored for reasoning VLAs. By leveraging a strong VLM as an assessor, {\evalname} provides fine-grained evaluation across reasoning, intention, action, and reasoning–action alignment, offering deeper insight into model behavior and failure modes.
    \item    
We conduct extensive experiments across simulation and real-world tasks, showing that {\modelname} consistently mitigates action degeneration and improves overall VLA performance. {\evalname} further reveals important bottlenecks in current VLA development and validates the strengths of our approach.
    
\end{itemize}

%% file: sec/2_related_work.tex
\section{Related Work}
\label{sec:related_work}
\input{tables/evolution_vla}

\paragraph{Vision-Language-Action(VLA) Models.} 
Leveraging the rich visual priors and zero-shot generalization capabilities of VLMs, VLAs fine-tune VLMs on robotic manipulation datasets~\cite{oxe-dataset,bridgev2,droid} to produce action signals, emerging as a scalable and promising direction in embodied intelligence~\cite{zhong2025survey,openvla,rt2,shukor2025smolvla}. OpenVLA~\cite{openvla} exemplifies this paradigm by finetuning Prismatic-7B~\cite{karamcheti2024prismatic} and mapping discrete tokens to action bins, while subsequent studies further explore architectural designs and action modeling~\cite{hybridvla,chen2025fast,pi0,mla,liu2025trivla,Li_2025_CVPR,wang2025vla,li2025cronusvla}. However, such specialist VLAs typically sacrifice general multimodal and reasoning capabilities, motivating the development of reasoning VLAs. By distilling reasoning traces from advanced VLMs, recent works construct high-quality embodied CoT data for supervised finetuning~\cite{ecot,emma-x,refinevla}; for example, Emma-X~\cite{emma-x} employs Gemini~\cite{gemini} to generate embodied CoT annotations for finetuning OpenVLA~\cite{openvla}. Incorporating multimodal corpora further enhances general understanding and reasoning~\cite{thinkact,molmoact,instructvla}. Nonetheless, compared with pre-finetuned specialist VLAs, these reasoning VLAs consistently exhibit degraded action performance, indicating catastrophic forgetting not only in the VLM-to-VLA transition but also in the specialist-to-generalist adaptation—ultimately hindering the development of truly generalizable embodied agents.

\paragraph{MLLM as a judge.} 
In natural language processing, the LLM-as-a-Judge paradigm~\cite{llm-as-a-judge} leverages the strong priors and zero-shot generalization of large language models to evaluate model outputs through contextual prompts~\cite{gu2024survey}, achieving notable success in mathematical reasoning~\cite{stephan2024calculation,zhang2025lessons}, function calling~\cite{toolbench,critictool}, agentic tasks~\cite{zhuge2024agent}, and other domains~\cite{chen2024not}. With the rise of R1-style models, assessing a model’s generated reasoning has become increasingly important~\cite{nguyen2024direct,jiang2025mme}. This motivates extending the paradigm to multimodal settings, which are inherently more complex and difficult to evaluate. Recent advances in VLMs~\cite{zeng2025agentic,wang2025vrag, wang-etal-2025-vidorag,zeng-etal-2025-enhancing-large,qi2025vcr,zhao2025v2p} and early explorations of MLLM-as-a-Judge~\cite{mllmasajudge} demonstrate its feasibility, further supported by systems like Llava-critic~\cite{xiong2025llava}, which provide both scores and rationales. MLLM-based evaluation has also shown robustness in tasks such as spatial reasoning~\cite{jia2025omnispatial} and controllable image generation~\cite{chen2025edit, huang2024vbench,huang2025interleaving}.
In contrast, existing VLA evaluations rely heavily on task success rates~\cite{calvin,libero,simplerenv}, overlooking reasoning quality and failing to capture beneficial trial-and-error behaviors. To address this gap, we introduce the MLLM-as-a-Judge paradigm into VLA evaluation. As VLMs continue to advance, we believe they will play an increasingly important role in evaluating VLA systems.
\begin{figure}[t]
    \centering
   
    \includegraphics[width=0.95\columnwidth]{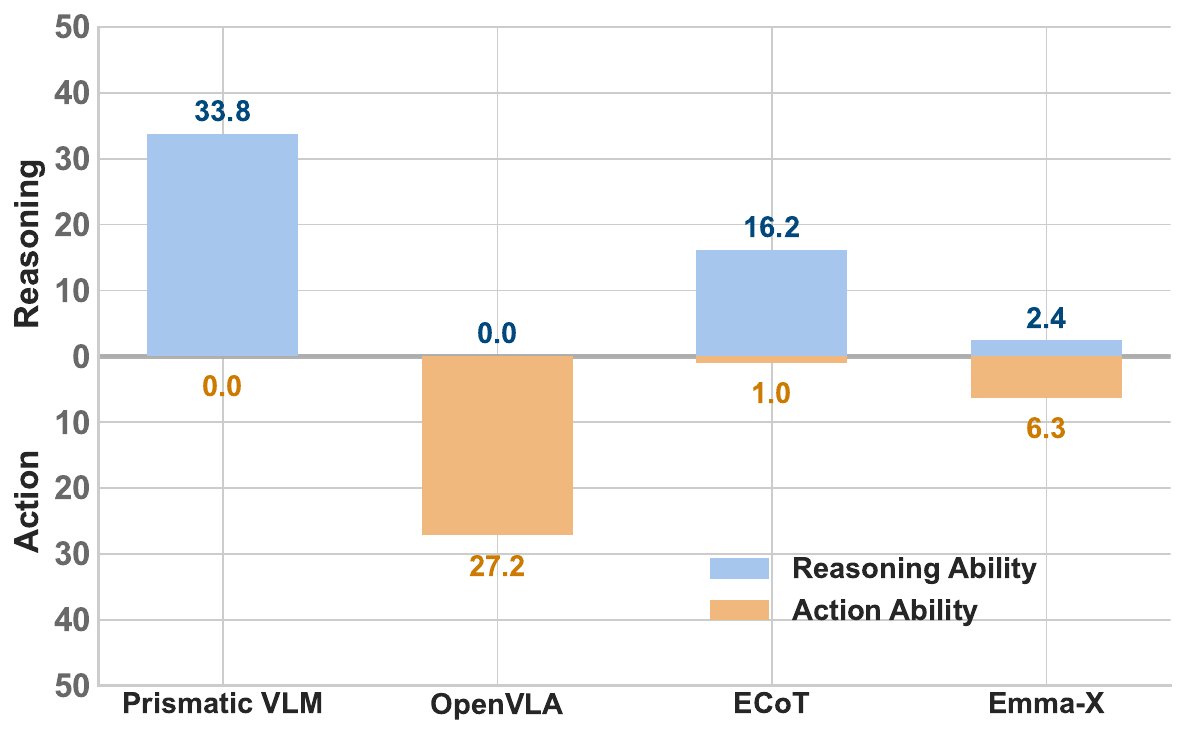}
  \caption{ \label{fig: action_decay} 
VLMs possess strong reasoning ability but lack action skills. Specialist VLAs achieve strong action capability but lose general reasoning. Reasoning VLAs partially recover reasoning through additional supervision, yet their action performance drops, illustrating the action degeneration problem. Our goal is to build a model that excels at both reasoning and action simultaneously.}
\vspace{-4mm}
\end{figure}

%% file: tables/evolution_vla.tex
\begin{table}[t]
\centering
\small
\caption{Comparison of base model, data, reasoning capability, and action capability from VLMs to specialist VLAs and finally to reasoning VLAs. SVLAs denotes specialist VLAs and RVLAs denotes reasoning VLAs.}

\label{tab: evolution_VLA}
\scalebox{1}{
    \begin{tabular}{c|c c c c}
    \toprule
     &  \textbf{Base Model} & \textbf{Data} & \textbf{Reason} & \textbf{Act} \\ 
     \midrule
    VLMs & LLMs & Corpora &  ↑ & -\\
    SVLAs & VLMs & Robot Data  & ↓ & ↑\\
    RVLAs & SVLAs & Both  &↑& \textbf{\textcolor{red}{↓}}  \\
    \bottomrule
    \end{tabular}
    }
\vspace{-1em}
\end{table}

%% file: sec/4_methd.tex
\section{Method}
In this section, we begin with the preliminaries of VLAs in Sec.~\ref{sec:pre}. Then we introduce {\modelname}, which combines dual-layer pruning in Sec.~\ref{sec:data} and dual-teacher distillation in Sec.~\ref{sec:teacher} to balance reasoning and action learning. Finally, we present {\evalname} in Sec.~\ref{sec:VLAs score}, a fine-grained evaluation framework for assessing both capabilities.

\subsection{Preliminaries}
\label{sec:pre}
Given a robotic dataset~\cite{bridgev2} of expert trajectories
\begin{equation}
    \mathcal{D}=\{\tau_i\}_{i=1}^{N}, \qquad
    \tau_i=\{(o_t,i_t,a_t)\}_{t=1}^{T_i},
\end{equation}
where \(o_t\) is the visual observation, \(i_t\) the instruction, and \(a_t\in\mathbb{R}^d\) the action.
A specialist VLA \(\pi_\theta(a\mid o,i)\) is trained via the autoregressive action loss:
\begin{equation}
\label{eq: s1}
\mathcal{L}_{\text{s}}(\theta)
= -\sum_{\tau_i}\sum_{t}\sum_{k}
\log \pi_\theta(a_{t,k}\mid o_t,i_t,a_{t,1:k-1}),
\end{equation}
where \(a_{t,1:k-1}\) are previously generated tokens.

For reasoning VLAs, each trajectory is augmented with reasoning \((o_t,i_t,r_t,a_t)\).
Multimodal samples share the unified form \((o,i,r,\emptyset)\), and the policy
\(\pi_\theta(r,a\mid o,i)\) jointly predicts reasoning and actions using:
\begin{equation}
\label{eq: s2}
\mathcal{L}_{\text{g}}(\theta)
= -\sum_{\tau_i}\sum_{t}\sum_{k}
\log \pi_\theta(y_{t,k}\mid o_t,i_t,y_{t,1:k-1}),
\end{equation}
where \(y_t\) denotes the joint reasoning–action sequence.

\subsection{{\modelname}}
A general embodied agent should exhibit both strong multimodal reasoning and precise action prediction. The emergence of {\decay} directly contradicts this goal. Our analyses indicate that this degradation arises from two factors:
(1) redundant, low-entropy embodied reasoning, which dominates the training sequence and interferes with visuomotor learning, and
(2) the absence of fine-grained, discriminative supervision needed to separate reasoning learning from action learning.
These observations highlight the need to decouple the two processes and provide differentiated supervision.

To address {\decay}, we propose {\modelname}. Motivated by the negative impact of overfitted embodied reasoning, we introduce a two-stage pruning strategy that removes redundant reasoning while preserving action-relevant segments. To directly enhance action capability, we further adopt a dual-teacher framework: an action teacher offers fine-grained supervision for robot data, while a multimodal teacher maintains general reasoning ability. Together, the two teachers provide balanced, task-specific guidance that jointly improves action execution and multimodal understanding.
\subsubsection{Dual-Layers Data Pruning}
\label{sec:data}
%
Embodied reasoning is often treated as an implicit curriculum that enables the model to learn high-level reasoning before acquiring fine-grained action skills~\cite{ecot-lite}. However, the reasoning tokens in embodied datasets are typically low-entropy, repetitive, and weakly coupled to the underlying visuomotor dynamics. This occurs because embodied scenarios (e.g., kitchen tasks) have limited object and action diversity (e.g., grasp, move, place), causing nearly identical reasoning text to be repeated across long temporal spans. During joint training, these abundant and homogeneous reasoning tokens dominate the loss and lead the model to overfit the linguistic reasoning patterns rather than learning the essential visuomotor relationships for manipulation. As a result, the action-relevant gradients are diluted or overridden, ultimately degrading the model’s action execution capability.
To mitigate this undesirable supervision bias, we selectively prune redundant reasoning and retain only segments that are truly action-critical. Using advanced VLMs to annotate such segments is costly and difficult to scale~\cite{onetwovla}, while relying solely on robotic kinematics ignores scene semantics and suffers from robustness issues~\cite{notvla}. Our dual-layer pruning addresses both challenges by jointly considering scene event changes and action dynamics.

Inspired by perception–action coupling theory~\cite{warren1990perception, farrow2003expertise}, which posits that cognitive effort is allocated only when action demands and environmental changes jointly require additional reasoning, we retain only reasoning contents with high reasoning value by video event boundary detection and kinematic key-frame selection.
For each frame, we assign a reasoning scene label and a reasoning action label. Keyframes are identified from two perspectives: video event changes and robot motion changes. Frames for which both labels are set to 1 are considered essential for reasoning.
For video event boundary detection, we manually annotate a small set of trajectories with reasoning scene labels, indicating whether each step requires reasoning.
Using high-quality annotated data, we retrained DDM-Net~\cite{Tang_2022_CVPR}, which is designed for generic event boundary detection.
Based on this well-pretrained detection net as the reasoning trigger, we obtain reasoning scene labels.
For kinematic key-frame selection, we focus on moments where abrupt velocity changes or gripper state transitions occur.
Let the end-effector pose be 
$T(t) = [x(t), y(t), z(t), \theta_x(t), \theta_y(t), \theta_z(t)]^\top$
and the gripper state be \( G(t) \in \{0,1\} \).
The reasoning action label of the keyframe at time \( t_k \) will be set to 1 if:
\begin{figure*}[ht]
    \centering
    \includegraphics[width=\textwidth]{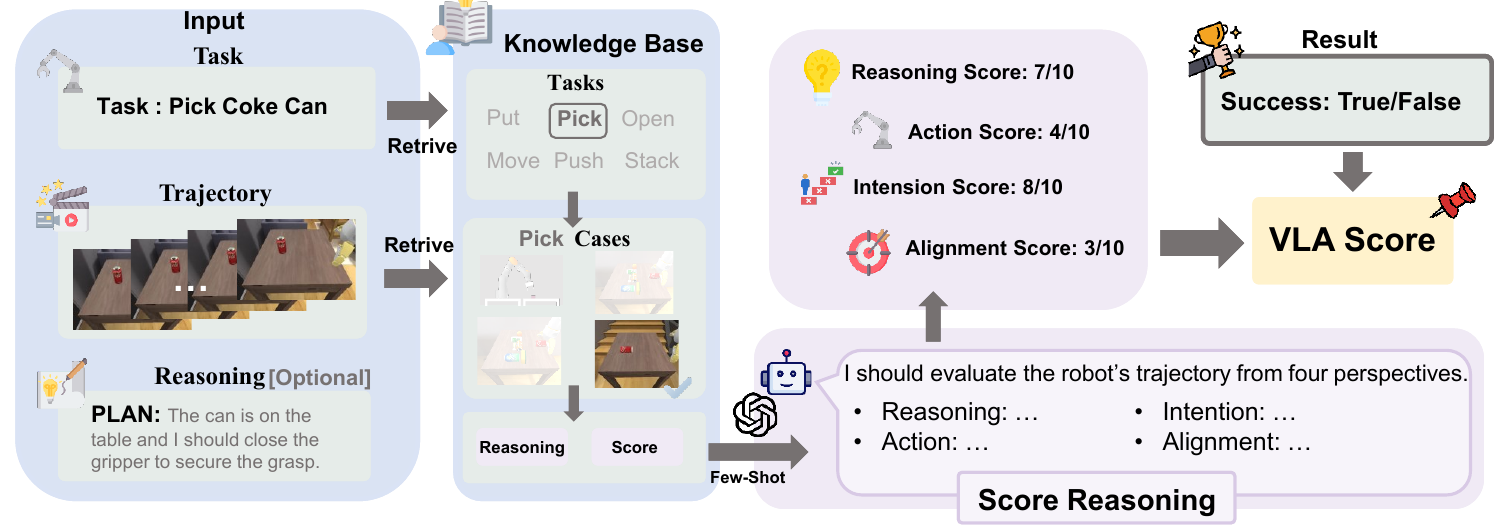}
      \vspace{-2em}
  \caption{\label{fig:eval_framework}\textbf{
Overview of {\evalname}  evaluation pipeline.} Given the policy trajectory, task description, and optional reasoning as input, \evalname{} first performs dual retrieval to fetch task-relevant textual examples and visually similar trajectories from a curated knowledge base. The retrieved samples serve as few-shot context for the VLM judge, which evaluates the trajectory along four dimensions: Reasoning, Action, Intention, and Alignment. These scores are then combined with the simulation outcome to produce the final {\evalname}.}
          \vspace{-1.5em}
\end{figure*}
\begin{equation}
\begin{split}
& \left\| \ddot{T}(t_k) \right\|_2 \;>\; \frac{1}{N} \sum_{i=1}^{N} \left\| \ddot{T}(t_i) \right\|_2 \\
&\lor \lim_{\epsilon \to 0^+} G(t_k-\epsilon) \neq \lim_{\epsilon \to 0^+} G(t_k+\epsilon)
\end{split}
\end{equation}
This approach is parameter-free compared with NoTVLA~\cite{notvla}.
We mark frames whose scene and action labels are both 1 as keyframes, retain their reasoning contents, and mask all others.
By jointly using scene-boundary detection and action-change cues, we prune redundant low-entropy reasoning and build a sparse, information-dense dataset.
This selective retention naturally forms a curriculum: the model first consolidates action grounding, then learns reasoning only when necessary, preventing early overfitting and mitigating {\decay}.

\subsubsection{Dual-Teacher Adaptive Distillation}
\label{sec:teacher}
While Sec.~\ref{sec:data} focuses on suppressing the negative effect of embodied-reasoning overfitting on action behavior, an orthogonal direction is to ask: can we directly improve the model’s action capability rather than merely preventing degradation? A key observation is that action degeneration fundamentally results from misaligned supervision signals: during mixed-modality finetuning, action tokens receive weak or noisy gradients, while reasoning tokens dominate the objective, leading the model to favor linguistic correctness over precise motor control.
Inspired by the proverb `Turn inward and examine yourself when you encounter difficulties in life', we revisit the source model from which the reasoning VLA is finetuned. Since specialist VLAs inherently possess strong action execution capabilities, their output distributions naturally provide fine-grained, action-aligned supervision that the reasoning VLA can benefit from. Thus, we introduce an action distillation loss that uses the specialist VLA as an action teacher to provide smooth, structured supervision over robot-centric samples:
\begin{equation}
\label{eq:action_kd_seq}
\mathcal{L}_{\text{action\_KD}}
=  T^2 D_{\mathrm{KL}}\!\bigl(\pi_{\theta_a}(a\mid o,i)\;\big\|\;\pi_{\theta}(a\mid o,i,r)\bigr).
\end{equation}
where $T>0$ is the temperature parameter used to soften the probability distributions and $\pi_{\theta_a}$ denotes the action teacher.

However, directly applying action distillation throughout training would cause the model to inherit the specialist VLA’s limited multimodal understanding ability. This reveals a second supervision-signal mismatch: while robot data require action-aligned supervision, multimodal reasoning data require reasoning-aligned supervision. To preserve multimodal competence, we therefore designate the finetuning initialization, which is already strong in general reasoning, as a reasoning teacher and introduce an analogous distillation objective:
%
\begin{equation}
\label{eq:action_kd_seq}
\mathcal{L}_{\text{reason\_KD}}
=  T^2 D_{\mathrm{KL}}\!\bigl(\pi_{\theta_r}(r\mid o,i)\;\big\|\;\pi_{\theta}(r\mid o,i)\bigr).
\end{equation}
where $\pi_{\theta_r}$ denotes the reasoning teacher.
Applying both teachers to all samples would be computationally expensive and exacerbate conflicts between heterogeneous supervision signals. Instead, the mixed-modality training regime naturally reveals which teacher to use: robot data receive action-aligned supervision, and multimodal reasoning data receive reasoning-aligned supervision. The final loss is:
%
%

\begin{equation}
\mathcal{L}_{\text{total}} = \mathcal{L}_{\text{VLA}} + \lambda\mathcal{L}_{KD}
\end{equation}
\[
\text{where } \mathcal{L}_{kd} = 
\begin{cases}
\mathcal{L}_{\text{action\_KD}}, & \text{if robot data} \\
\mathcal{L}_{\text{reason\_KD}}, & \text{if multimodal data}
\end{cases}
\]
$\mathcal{L}_{\text{VLA}}$ corresponds to the VLA training loss using hard-label cross-entropy and  $\lambda$ is the  auxiliary weight and set to 0.15.
By constructing this isomorphic dual-teacher adaptive distillation strategy and assigning supervision signals that match the nature of each data modality, the model learns both reasoning and action under fine-grained, aligned guidance. Compared to hard labels, soft supervision reduces gradient conflicts and restores the appropriate balance between linguistic reasoning and low-level control, thereby alleviating action decay and leading to a more generalizable embodied agent.
%
\subsection{{\evalname}}
\label{sec:VLAs score}

At present, VLAs evaluation primarily relies on the success rate. However, this sparse and coarse-grained metric cannot capture critical aspects of VLAs performance, such as action smoothness, reasoning correctness, and the degree to which actions adhere to the inferred plan.
To address this gap, we propose {\evalname}, the first metric specifically designed for fine-grained VLAs evaluation.
%
%
It is worth noting that {\evalname} remains compatible with VLAs models lacking explicit reasoning capabilities and can be seamlessly adapted to various simulation environments.
Fig.~\ref{fig:eval_framework} shows the pipeline of \evalname.
Given the trajectory of policy $\pi_\theta$, task description, reasoning content (if available), and simulation success indicator, we prompt GPT-4o~\cite{gpt4o} to perform evaluations from four perspectives as follows:
\begin{itemize}
    \item \textbf{Reasoning Score} $R$: Measures the correctness, logical consistency, and usefulness of the reasoning process in guiding the agent toward successful task completion.
    \item \textbf{Action Score} $A$: 
    Measures the coherence and smoothness of the action sequence.
    \item \textbf{Intention Score} $I$: Determines whether the model’s actions contribute constructively to solving the task.
    \item \textbf{Reason–Act Alignment Score} $RA$: Measures how well the action sequences align with reasoning content.

\end{itemize}
Combined with the trajectory’s simulation result $B$, we calculate the overall VLAs Score using the following formula:
\input{tables/simplerenv4}
{\small
\begin{equation}
\evalname =
\begin{cases}
\left( \dfrac{R + A \cdot I}{2} \right) \cdot RA \cdot B, & \pi_\theta \in \text{RVLA} \\[4pt]
A \cdot I \cdot RA \cdot B, & \pi_\theta \notin \text{RVLA}
\end{cases}
\end{equation}
\[
\text{where } B = 
\begin{cases}
1, & \text{if success} \\
0, & \text{if fall}
\end{cases}
\]
}

To improve evaluation accuracy and generalization, we adopt a retrieval-enhanced judge. We first collect 100 diverse trajectories and obtain preliminary scores using the previous pipeline. Human experts refine these results to form a VLAs knowledge base, which the judge retrieves during evaluation.
For each input trajectory, we apply a dual-retrieval mechanism: task retrieval encodes the textual description and retrieves the most relevant task from the knowledge base, and scene retrieval encodes the image frames to fetch the most similar annotated trajectory as contextual reference. This retrieved context improves the VLM's evaluation accuracy. 
We use text-embedding-ada-002~\footnote{https://platform.openai.com/docs/models/text-embedding-ada-002} as the text embedding encoder and CLIP ViT-B/32~\cite{clip} as the image embedding encoder, both of which are commonly adopted in Retrieval-Augmented Generation (RAG) pipelines.


%% file: tables/simplerenv4.tex
\begin{table*}[t]
\small
\centering
\vspace{3pt}
\caption{\textbf{Comparison of manipulation success rates between {\modelname} and specialist \& generalist baselines in SimplerEnv.} 
Google Robot and WidowX Robot denote two embodiments in SimplerEnv. VM refers to visual matching and VA refers to variance aggregation. 
$^{\dagger}$ denotes models without released checkpoints, results are taken from their papers.}
\label{tab: manip}
{
\begin{tabular*}{0.9\textwidth}{l @{\extracolsep{\fill}} cccccccccc}
\toprule
\multirow{3}{*}{\textbf{Methods}} &
\multicolumn{6}{c}{\textbf{Google Robot}} &
\multicolumn{3}{c}{\textbf{WidowX Robot}} &
\multirow{3}{*}{\textbf{Avg}}
\\
\cmidrule(lr){2-10}
 & 
 \multicolumn{2}{c}{Drawer} &
 \multicolumn{2}{c}{Pick Can} &
 \multicolumn{2}{c}{Move Near} &
 \multirow{2}{*}{\shortstack{Put \\ Spoon}} & 
 \multirow{2}{*}{\shortstack{Put \\ Carrot}} & 
 \multirow{2}{*}{\shortstack{Stack \\ Blocks}}
 &
\\
\cmidrule(lr){2-7}
 & VM & VA & VM & VA & VM & VA & & & & \\ 
\midrule

RT-1-X~\cite{oxe-dataset}
& 59.7 & 29.4 & 56.7 & 49.0 & 31.7 & 32.3 & 0.0 & 4.2 & 0.0 & 26.8 \\

RT-2-X~\cite{oxe-dataset}
& 25.0 & 35.5 & 78.7 & 82.3 & 77.9 & 79.2 & - & - & - & - \\

Octo-Base~\cite{octo}
& 22.7 & 1.1 & 17.0 & 0.6 & 4.2 & 3.1 & 15.8 & 12.5 & 0.0 & 7.0 \\

RoboVLMs~\cite{robovlms}
& 43.5 & 10.6 & 77.3 & 75.6 & 61.7 & 60.0 & 45.8 & 20.8 & 4.2 & 38.8 \\

TraceVLA~\cite{tracevla}
& 63.1 & 61.6 & 45.0 & 64.3 & 63.8 & 60.6 & 12.5 & 16.6 & 16.6 & 38.6 \\

OpenVLA~\cite{openvla}
& 63.0 & 28.8 & 18.0 & 60.8 & 56.3 & 67.7 & 4.2 & 0.0 & 0.0 & 27.2 \\

SpatialVLA~\cite{spatialvla}
& 57.4 & 41.8 & 86.0 & 88.0 & 77.9 & 72.7 & 16.7 & 25.0 & 29.2 & 45.9 \\

InstructVLA-E~\cite{instructvla}
& 47.2 & 60.6 & 87.7 & 76.0 & 68.3 & 77.3 & 45.8 & 20.8 & 20.5 & 56.0 \\

\midrule
Magma~\cite{magma}
& 9.7 & 5.8 & 46.0 & 46.4 & 60.0 & 82.0 & 45.8 & 33.3 & 8.3 & 30.5 \\

ECoT~\cite{ecot}
& 0.0 & 0.0 & 0.0 & 0.0 & 0.2 & 0.2 & 4.2 & 4.2 & 0.0 & 1.0 \\

Emma-X~\cite{emma-x}
& 18.3 & 20.5 & 2.3 & 5.3 & 3.3 & 7.3 & 0.0 & 0.0 & 0.0 & 6.3 \\

ThinkACT$^{\dagger}$~\cite{thinkact}
& 50.0 & 47.6 & 92.0 & 84.0 & 72.4 & 63.8 & 58.3 & 37.5 & 8.7 & 57.1 \\

InstructVLA-G~\cite{instructvla}
& 55.0 & 55.2 & 77.2 & 90.8 & 54.1 & 70.0 & 33.3 & 29.2 & 12.5 & 53.0 \\

\midrule
\textbf{\modelname}
& 63.9 & 64.0 & 93.3 & 82.7 & 60.9 & 75.3 & 50.0 & 50.0 & 8.3 & \textbf{61.0} \\
\bottomrule
\end{tabular*}
}
\vspace{-1em}
\end{table*}

%% file: sec/5_experiments.tex
\section{Experiments}

\begin{figure*}[ht]
    \centering
    \includegraphics[width=\textwidth]{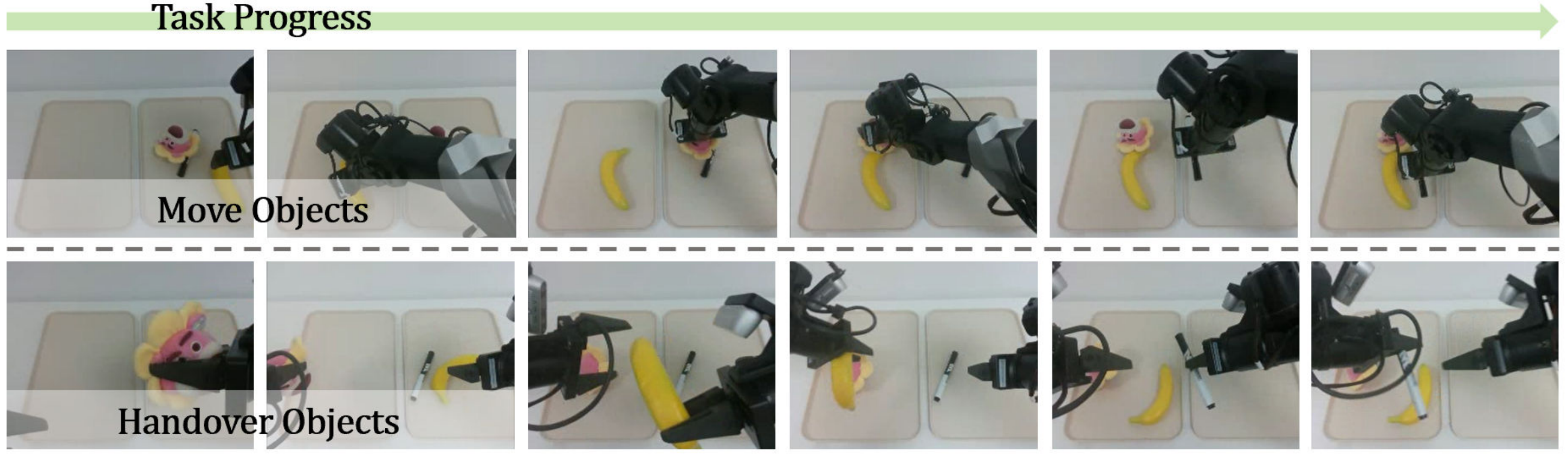}
      \vspace{-2em}
  \caption{\label{fig:real_progress} \textbf{Visualization} of the two real-world task progress.}
          \vspace{-1em}
\end{figure*}

\subsection{Main Results}
%
\paragraph{Implementation details.}
We first evaluate on SimplerEnv~\cite{simplerenv}, a widely adopted benchmark in the robotics community. By configuring visual matching and variance aggregation, SimplerEnv enables a more reliable assessment. 
We reconstruct the 650k VLA-IT~\cite{instructvla} dataset following the procedure described in Sec~\ref{sec:data} and subsequently fine-tune InstructVLA-G~\cite{instructvla} on the reconstructed dataset. 
We designate InstructVLA-E as the action teacher and InstructVLA-G as the multimodal reasoning teacher.
%
The learning rate is fixed at 2e-5.
%
{\modelname} addresses {\decay} that emerges when transitioning from specialist VLAs to reasoning VLAs. To prove it, both types of VLAs are included in the comparison.: \textbf{(1) specialist VLAs},  including RT-1-X and RT-2-X
from OXE~\cite{oxe-dataset}, Octo~\cite{octo}, RoboVLMs~\cite{robovlms}, SpatialVLA~\cite{spatialvla}, TraceVLA~\cite{tracevla}, InstructVLA-E~\cite{instructvla} and OpenVLA~\cite{openvla}.
(2) \textbf{reasoning VLAs}, including ECoT~\cite{ecot}, Emma-X~\cite{emma-x}, Magma~\cite{magma}, ThinkACT~\cite{thinkact}, and InstructVLA-G~\cite{instructvla}.

\paragraph{Quantitative Results.} In Tab.~\ref{tab: manip}, {\modelname} achieves an average success rate of 61.0 on the SimplerEnv benchmark.
{\modelname} improves upon the baseline InstructVLA-G by 8.0 average success rate.
This indicates that {\modelname} successfully resolved {\decay}.
Compared with other prior methods, {\modelname} improves the mean success
rate by 5.0 over the top-performing specialist VLA InstructVLA-E method and by 3.9 over the report result of top-performing reasoning VLA ThinkACT.
Moreover, we observe an emergent teacher surpassing phenomenon, where {\modelname} achieves higher scores than the action teacher.
And our pruning strategy yields about a 20\% inference speedup over the baseline.
Experimental results demonstrate that the model consistently improves performance across multiple tasks and settings, indicating that our carefully designed data pruning and dual-teacher adaptive distillation strategy effectively resolves the action decay phenomenon and further outperforms specialist VLAs.

\subsection{Real-world Results}

\paragraph{Real-world Setup.} To systematically evaluate our approach, we designed two real-world dual-arm tasks on the Galaxea R1-lite robot. For the dual-arm setting, three RealSense 455 camera are used to get image observations, one on the head and two on the left and right wrist, respectively. The model takes the images of three views as image observation, and outputs a 14-DoF vector as the dual-arm action.
\paragraph{Self-collected Data.} We designe two complex tasks:(1) Move Objects, (2) Handover Objects. Both tasks require the model to move three objects from right to left and follow the order in the language instruction. For each task, we collected 50 high-quality demonstration trajectories. Fig.~\ref{fig:real_progress} shows the progress of the two tasks.
\input{tables/real_world}
\paragraph{Quantitative Results.}
We test 10 rollouts for each task and use the average success rate as the quantitative result. Tab.~\ref{tab:real_task} shows that {\modelname} significantly improves manipulation performance, raising the average success rate from 45\% to 60.0\% in real-world tasks.
The gains in both Move and Handover tasks demonstrate more reliable and coordinated action generation in real robotic settings.

\subsection{{\evalname} Results}
\paragraph{Can VLM evaluate VLA?}
Before analyzing {\evalname}, a natural question should be clarified: can GPT-4o evaluate VLA capabilities? 
{\evalname} evaluation can be formulated as an extension of video understanding. It involves recognizing objects, inferring required actions, determining whether the action and reasoning are consistent, and assessing whether the robot arm moves toward the target. Extensive prior work~\cite{arnab2025temporal,li2025advancing,zhang2024omagen,li2025self, Li_2023_ICCV} has demonstrated that VLMs possess strong video understanding abilities.
%
%
\paragraph{Quantitative Results. }
In Tab.~\ref{tab:vla-score-specialist} and Tab.~\ref{tab:vla-score-generalist}, {\modelname} achieves the highest score in {\evalname} in reasoning VLAs.
We observe that for specialist VLAs, the intention score is consistently slightly higher than the action score, indicating that the main bottleneck in their development lies in achieving smoother, more robust, and more efficient action modeling.
For reasoning VLAs, the reasoning score is significantly higher than both the action and alignment scores. From analyzing failure cases, we find that the low action score primarily results from the model’s inability to execute effective actions throughout the trajectory. Even when it correctly reasons how to do, it often struggles to approach or manipulate the target.
{\modelname} inherits the reasoning ability of the reasoning teacher while acquiring the teacher model’s more refined and smooth actions, which cross-entropy alone cannot achieve. We posit that this combination is a key factor behind {\modelname}’s effectiveness.
%
\input{tables/vla_score_result}

\subsection{Ablation Study}
%

\begin{figure}[htbp]
  \centering
  
  \begin{minipage}[t]{0.48\linewidth}
    \centering
    \includegraphics[width=\linewidth]{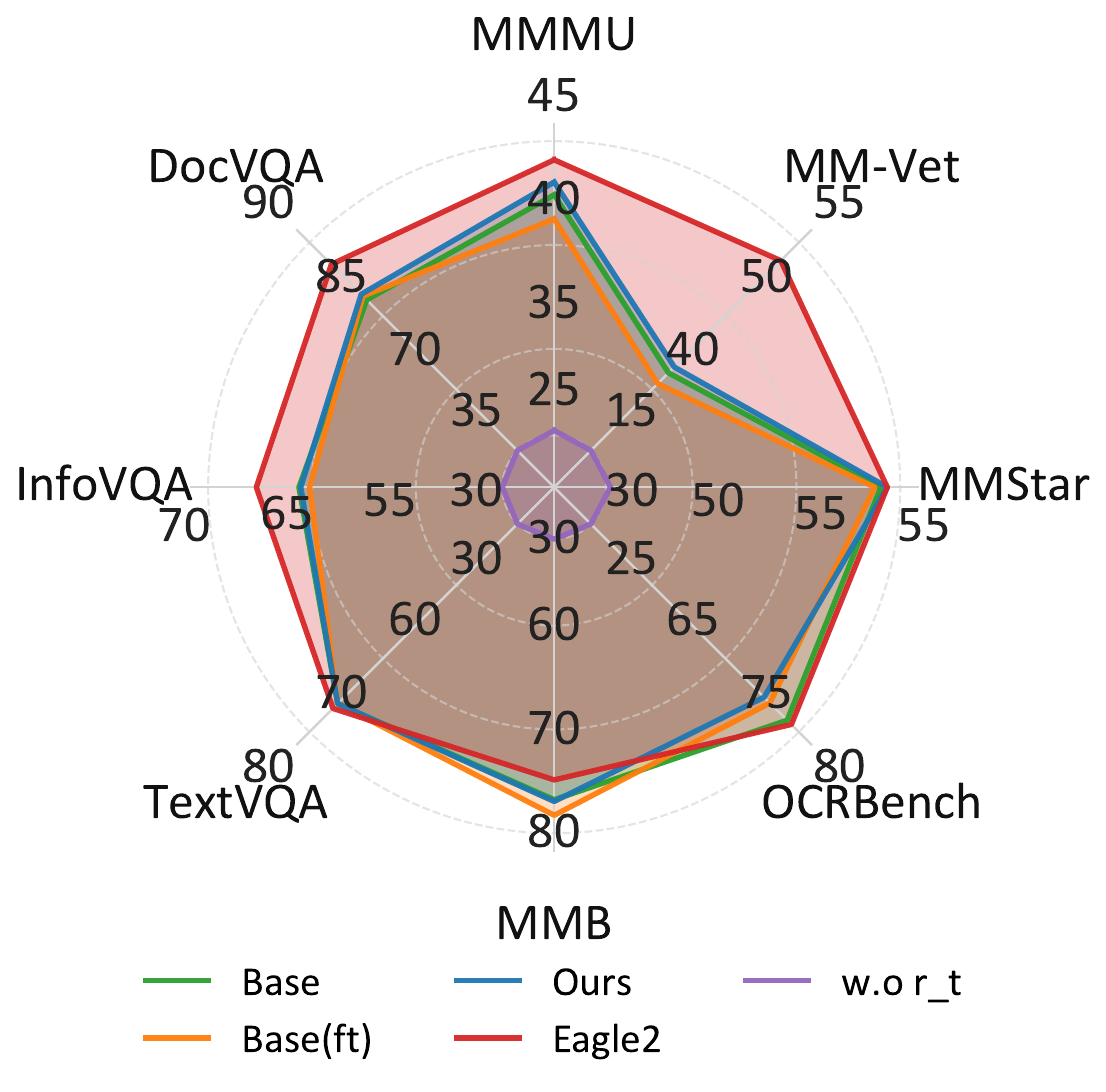}
    \caption{Ablation for distillation.}
    \label{fig:ablation_mm_teacher}
  \end{minipage}
  \hfill
  \begin{minipage}[t]{0.44\linewidth}
    \centering
    \includegraphics[width=\linewidth]{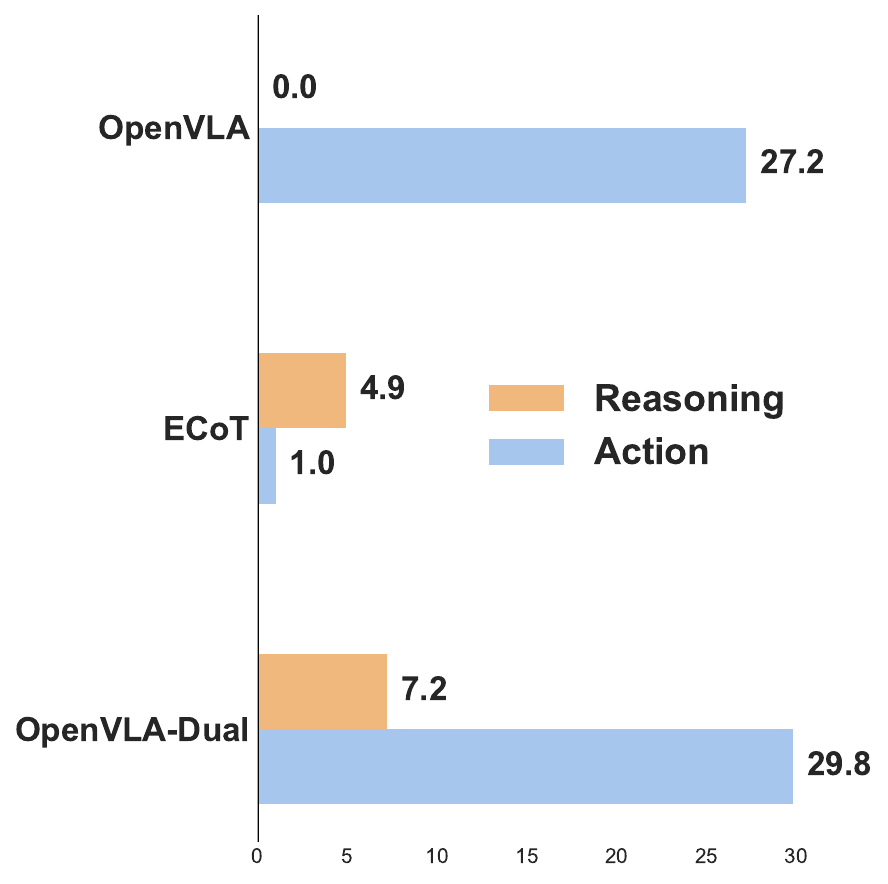}
    \caption{Ablation for base models.}
    \label{fig:ablation_model}
  \end{minipage}
  \vspace{-3em}
\end{figure}

We still employ SimplerEnv for robotic manipulation evaluation, and assess multimodal tasks on MMMU~\cite{MMMU}, MM-Vet~\cite{MMVET}, MMStar~\cite{MMSTAR}, OCRBench~\cite{Ocrbench}, MMB~\cite{MMB}, TextVQA~\cite{TextVQA}, InfoVQA~\cite{infovqa} and DocVQA~\cite{DOCVQA} for all ablation studies. MM denotes the average of these multimodal understanding and QA benchmarks.
GR denotes Google Robot set and WR denotes WidowX Robot.
Owing to the limited default number of trials in WidowX (only 24 per task), \textbf{we repeat every ablation experiment five times in the main paper, yielding 120 trajectories per task.}
%
Following a coarse-to-fine principle, we first perform macro-level ablations to evaluate the overall contributions of data pruning and distillation. 
As shown in Tab.~\ref{tab:ablation_module}, Base denotes the baseline InstructVLA-G, and Base(ft) denotes fine-tuning without data pruning and distillation strategies.
Plain fine-tuning results in only marginal gains, showing diminishing returns.
Teacher distillation significantly improves action execution across tasks while preserving multimodal capability. Data distillation further boosts performance by reducing the adverse influence of excessive embodied reasoning.
\input{tables/abaltion_module}
\paragraph{Pruning and Distillation.}
\input{tables/ablation_data}
%
Tab.~\ref{tab:ablation_data} shows that two-layer data pruning achieves better performance than single-layer pruning and random dropout. 
Scene pruning contributes the majority of the performance gain, while action pruning provides an additional complementary improvement. Moreover, the double-layer pruning strategy consistently outperforms proportional random dropout, showing that its effectiveness stems from the deliberate pruning design rather than randomness like ECoT-Lite~\cite{ecot-lite}.
%
\input{tables/ablation_teacher}
For dual-teacher adaptive distillation, we primarily investigate whether the model can preserve its multimodal general capabilities while improving action performance.
As shown in Tab~\ref{tab:ablation_teacher}, 
the dual-teacher adaptive design enables the model to simultaneously strengthen action generation capability and preserve general multimodal understanding. 
Although using only the action teacher yields a slight improvement in action performance, it comes at the cost of a substantial drop in multimodal capability, highlighting that mere action-specific supervision cannot maintain generalization.
Fig.~\ref{fig:ablation_mm_teacher} shows that distillation from the reasoning teacher effectively preserves multimodal reasoning capabilities, with no significant difference compared to the reasoning teacher or the base VLM. This suggests that the current bottleneck in advancing reasoning VLAs lies in their action proficiency.
Full distillation from a single teacher for all domain data reduces to a plain Kullback-Leibler (KL) Divergence loss, offering no performance gain.

\paragraph{OpenVLA Variant}
To assess the extrapolation ability of {\modelname}, we extend our framework to OpenVLA~\cite{openvla} and ECoT~\cite{ecot}, yielding OpenVLA-Dual.
OpenVLA-Dual surpasses the action teacher (OpenVLA) in action performance and outperforms the reasoning teacher (ECoT) in multimodal generalization, confirming the practicality and broad applicability of our method.
Although the action gains are modest, we attribute this to the inherent constraints of OpenVLA’s discrete-token action representation, which limits the ceiling of motion optimization.


%

%% file: tables/real_world.tex
\begin{table}[!t]
\centering
\setlength{\tabcolsep}{3pt}
\small
\caption{Quantitative results for real-world experiments. We report the success rate of moving the 3 objects separately for the long-horizon case.  \textbf{Bold} indicates the highest score across all the models. }
\vspace{-4pt}
\begin{tabular}{l|ccc|ccc|c}
\toprule
\multirow{2}{*}{Method} 
  & \multicolumn{3}{c|}{Move} 
  & \multicolumn{3}{c|}{Handover} 
  & \multirow{2}{*}{Average}  
\\
\cmidrule(lr){2-4} \cmidrule(lr){5-7}
& O1 & O2 & O3 & O1 & O2 & O3 & 
\\
\midrule
InstructVLA~\cite{instructvla} & 60 & 60 & 50 & 60 & 40 & 40 & 45 \\
Ours & \textbf{90} & \textbf{80} & \textbf{70} & \textbf{80} & \textbf{60} & \textbf{50} & \textbf{60} \\
\bottomrule
\end{tabular}
\vspace{-8pt}
\label{tab:real_task}
\end{table}

%% file: tables/vla_score_result.tex




\begin{table}[tb]
\raggedright  
\centering
\small
\caption{Comparison between {\modelname} and specialist VLAs on {\evalname}. Sim. denotes the average success rate on SimplerEnv}
\label{tab:vla-score-specialist}
\begin{tabular}{lcccc}
\toprule
\textbf{Methods} &
\textbf{A} &
\textbf{I} &
\textbf{Sim.} &
\textbf{\evalname} \\ 
\midrule
RoboVLMs-2B~\cite{robovlms}&  62.0 & 71.8 &  38.8 & 31.2 \\
TraceVLA-7B~\cite{tracevla} & 53.3 & 62.4 & 38.6 & 23.2 \\
OpenVLA-7B~\cite{openvla}& 45.3 & 57.7  & 27.2 & 14.2 \\
SpatialVLA-3B~\cite{spatialvla} & 61.9 & 70.5 & 45.9 & 40.1 \\
InstructVLA-E~\cite{instructvla} &  \textbf{63.1} & \textbf{73.2} & \textbf{56.0} & \textbf{45.3} \\
\bottomrule
\end{tabular}
\end{table}
\begin{table}[tb]
\raggedright  
\centering
\footnotesize
\caption{Comparison between {\modelname} and reasoning VLAs.}
\label{tab:vla-score-generalist}
\begin{tabular}{lcccccc}
\toprule
\textbf{Methods} &
\textbf{R} &
\textbf{A} &
\textbf{I} &
\textbf{RA} &
\textbf{Sim.} &
\textbf{\evalname} \\ 
\midrule
ECoT~\cite{ecot}& 47.2 & 27.9 & 39.9 & 29.6 & 0.8 & 0.0 \\
Emma-X~\cite{emma-x} & 54.1 & 39.2 & 37.2 & 29.2 & 6.3 & 2.2 \\
Magma~\cite{magma} & 66.0 & 61.0 & 71.2 & 36.0 & 30.5 & 19.5 \\
Base~\cite{instructvla} & 72.3 & 48.6 & 68.0 & 43.3 & 53.0 & 35.1 \\
\midrule
\textbf{Ours}& \textbf{74.0} & \textbf{64.8} &\textbf{74.7} & \textbf{51.0} & \textbf{61.0} & \textbf{42.9} \\
\bottomrule
\end{tabular}
\vspace{-1em}
\end{table}

%% file: tables/abaltion_module.tex
\begin{table}[!t]
\centering
\small
\caption{Ablation of two strategies introduced in {\modelname}.
}
\vspace{-4pt}

\begin{tabular}{l|ccc|c}
\toprule
   Setting & GR & WR & MM & Average \\
\midrule
   Base     & 67.1 & 25.0 & \textbf{65.5}&52.5 \\ 
   Base(ft)   & 66.1 & \textbf{33.3} & 65.1&54.8\\ 
   w.o. Pruning   & 72.4 & 29.2 & 65.0&55.6 \\
  w.o. Distillation & 70.4 & 28.1 & 65.2&54.5 \\
    Ours    & \textbf{73.4} & 31.1 & 65.4 & \textbf{56.6}  \\
\bottomrule

\end{tabular}

\vspace{-8pt}
\label{tab:ablation_module}
\end{table}

%% file: tables/ablation_data.tex
\begin{table}[!t]
\centering
\small
\caption{Ablation for different purning methods.
}
\vspace{-4pt}
\begin{tabular}{l|ccc|c}
\toprule
   Setting & GR & WR & MM & Average \\
\midrule
     Ours    & \textbf{73.4} & \textbf{31.1} & \textbf{65.4} & \textbf{56.6} \\
Random    & 63.6 & 23.1 & 64.6&50.4 \\ 
   w.o. Action   & 72.0 & 30.3 & 64.8 & 55.7 \\
  w.o. Scene   & 66.8 & 28.3 & 61.3 & 52.1 \\

\bottomrule

\end{tabular}
\vspace{-8pt}
\label{tab:ablation_data}
\end{table}

%% file: tables/ablation_teacher.tex
\begin{table}[!t]
\centering
\small
\caption{Ablation for distillation strategies. a\_t denotes the action teacher and r\_t denotes the reasoning teacher.
}
\vspace{-4pt}
\begin{tabular}{l|ccc|c}
\toprule
   Setting & GR & WR & MM & Average \\
\midrule
     Ours    & 73.4 & 31.1 & \textbf{65.4} & \textbf{56.6} \\
w.o. a\_t    & 67.8 & 26.1 & 65.2& 53.0 \\
   w.o. r\_t   & \textbf{74.3} & \textbf{34.7} &30.7 & 46.6\\

\bottomrule

\end{tabular}
\vspace{-8pt}
\label{tab:ablation_teacher}
\end{table}

%% file: sec/6_conclusion.tex
\section{Conclusion, Limitations and Future Work}
In this paper, we propose {\modelname}, which mitigates action degeneration through dual-layer pruning and dual-teacher adaptive distillation, achieving stronger action execution while preserving reasoning. 
A limitation is the reliance on two teachers; although both are pretrained and introduce no extra training cost, this dependency still adds structural complexity.
Another limitation is the increased forward passes required during distillation; although forward computation is not the main bottleneck in VLA training, it still adds overhead, and our preliminary attempt at attention-level distillation shows that similar benefits can be retained with reduced computation. Future work will focus on simplifying the distillation pipeline, reducing the reliance on teacher models, and extending the framework to broader embodied settings.


%% file: sec/99_supp.tex
\clearpage
\appendix

\definecolor{lightgray}{gray}{0.95}
\definecolor{deepblue}{RGB}{70,130,180}
\definecolor{deepgray}{RGB}{119,136,153}
\lstdefinestyle{prompt}{
    basicstyle=\ttfamily\fontsize{7pt}{8pt}\selectfont,
    frame=none,
    breaklines=true,
    backgroundcolor=\color{lightgray},
    breakatwhitespace=true,
    breakindent=0pt,
    escapeinside={(*@}{@*)},
    numbers=none,
    numbersep=5pt,
    xleftmargin=5pt,
    aboveskip=2pt,
    belowskip=2pt,
}
\tcbset{
  aibox/.style={
    top=10pt,
    colback=white,
    enhanced,
    center,
  }
}
\newtcolorbox{AIbox}[2][]{aibox, title=#2,#1}

\section{\evalname}
The prompts are shown in Fig.~\ref{fig:reasoning-eval-prompt} and ~\ref{fig:specialist-eval-prompt}.

\section{Additional Qualitative Analyses}
\begin{figure}[t]
    \centering
    \includegraphics[width=0.85\columnwidth]{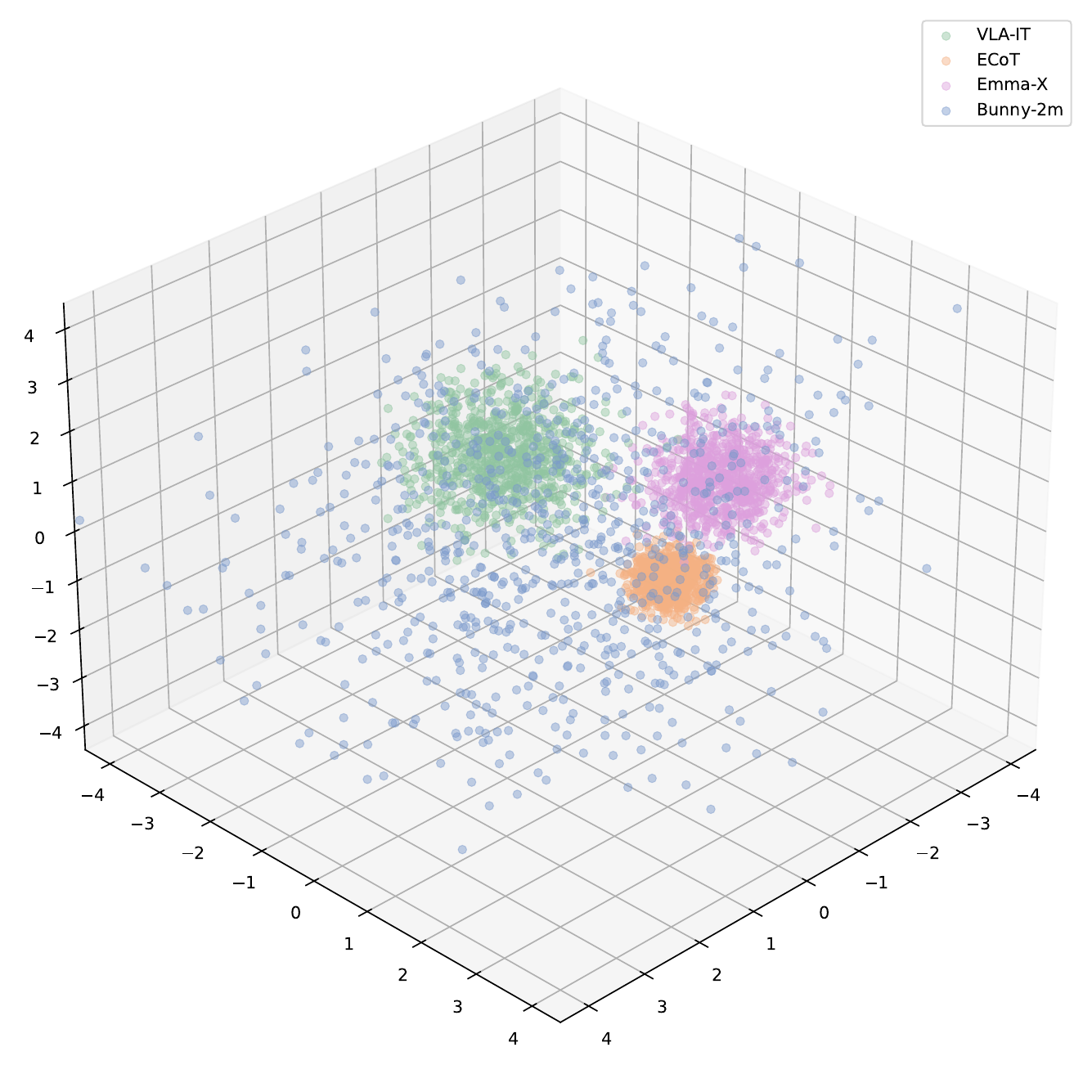}
  \caption{ \label{fig: data_vis} 
  \textbf{Embedding visualization of embodied reasoning and multimodal data.}
Robotic reasoning samples form dense, low-entropy clusters, whereas multimodal data remain more dispersed, indicating higher semantic diversity.
}
\vspace{-5mm}
\end{figure}

\begin{figure}[!t]
    \centering
    \includegraphics[width=0.95\columnwidth]{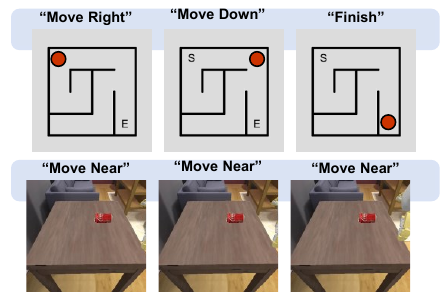}
  \caption{ \label{fig: data_example} 
\textbf{Example of redundant embodied reasoning across consecutive robot frames.}
Despite scene and motion changes, multiple frames share identical reasoning (e.g., “Move Near”), revealing redundancy that motivates pruning.}
\vspace{-5mm}
\end{figure}

\paragraph{Data Pruning.}
Fig.~\ref{fig: data_vis} visualizes the embodied reasoning embeddings from robot datasets. We observe that embodied reasoning in robotic datasets is highly redundant and densely clustered. As illustrated in Fig.~\ref{fig: data_example}, although the visual scenes continually change across an action sequence, the associated reasoning remains nearly identical. For example, when a robot approaches an object, multiple consecutive frames correspond to the same reasoning, such as “Move Near”, despite perceptible motion in the visual input. This redundancy arises from the nature of embodied tasks: trajectories often contain many low-level movement steps but only a small number of distinct semantic intentions. As a result, different frames map to nearly identical reasoning statements, creating a low-entropy and tightly clustered distribution. Such concentrated reasoning offers limited additional supervisory value yet dominates the training signal, thereby biasing the model toward overfitting trivial embodied reasoning rather than improving manipulation.
To mitigate the adverse effect of redundant embodied reasoning, we selectively prune repetitive samples rather than uniformly preserving the full corpus. Consider a training objective that combines action labels and reasoning supervision:
\begin{equation}
    \mathcal{L}=\mathbb{E}*{(x,y)\sim \mathcal{D}}\big[\ell{\text{act}}(x,y)+ \ell_{\text{reason}}(x,y)\big]
\end{equation}
When embodied trajectories contain highly repetitive states, the distribution $\mathcal{D}$ becomes skewed toward low-information reasoning segments, causing the expectation to overweigh $\ell_{\text{reason}}$ at these redundant points. In this scenario, minimizing $\mathcal{L}$ encourages the model to focus on predictable, low-entropy textual reasoning patterns instead of learning informative action–state relations. Consequently, the training bias suppresses the action objective $\ell_{\text{act}}$, degrading the model’s motion capability. Pruning concentrated, repetitive reasoning samples rebalances $\mathcal{D}$, enlarging the contribution of diverse state–action pairs and restoring meaningful supervision. Thus, pruning is not merely data reduction, but a necessary step to adjust the effective training distribution and preserve action learning.
\paragraph{Why does {\modelname} work?}
Training a reasoning VLA naturally involves optimizing two competing objectives, namely accurate action execution and robust multimodal reasoning. Let the corresponding goals be denoted as
$
\mathcal{L}{\text{act}}, \qquad \mathcal{L}{\text{reason}}.
$
Since their gradients usually point to different directions in parameter space, learning should not be viewed as minimizing a single scalar objective, but as converging to a Pareto optimal solution, where no capability can be further improved without degrading the other. In practice, naïvely mixing robot data with embodied reasoning drives optimization toward reasoning-dominant directions, because redundant and low-entropy reasoning traces generate disproportionately strong update signals. As a result, the model tends to converge toward the interior of the Pareto set, corresponding to suboptimal action performance despite moderate reasoning improvements.
To mitigate this imbalance, our method suppresses over-represented reasoning signals through pruning while enforcing structured gradient alignment via dual-teacher supervision. Pruning attenuates the dominance of
$
\nabla \mathcal{L}_{\text{reason}},
$
caused by repeated contextual traces, whereas expert-guided distillation maintains balanced gradients that avoid weakening
$
\nabla \mathcal{L}_{\text{act}}.
$
Together, these mechanisms steer optimization toward the Pareto-efficient frontier, yielding a VLA that enhances manipulation performance without sacrificing general multimodal capability. This geometric perspective explains why pruning or distillation alone is insufficient, while their combination effectively leads to a more desirable Pareto-efficient balance between action and reasoning.









\section{Case Study}
\subsection{Additional Visualization}
\begin{figure*}[t]
    \centering
    \includegraphics[width=0.95\textwidth]{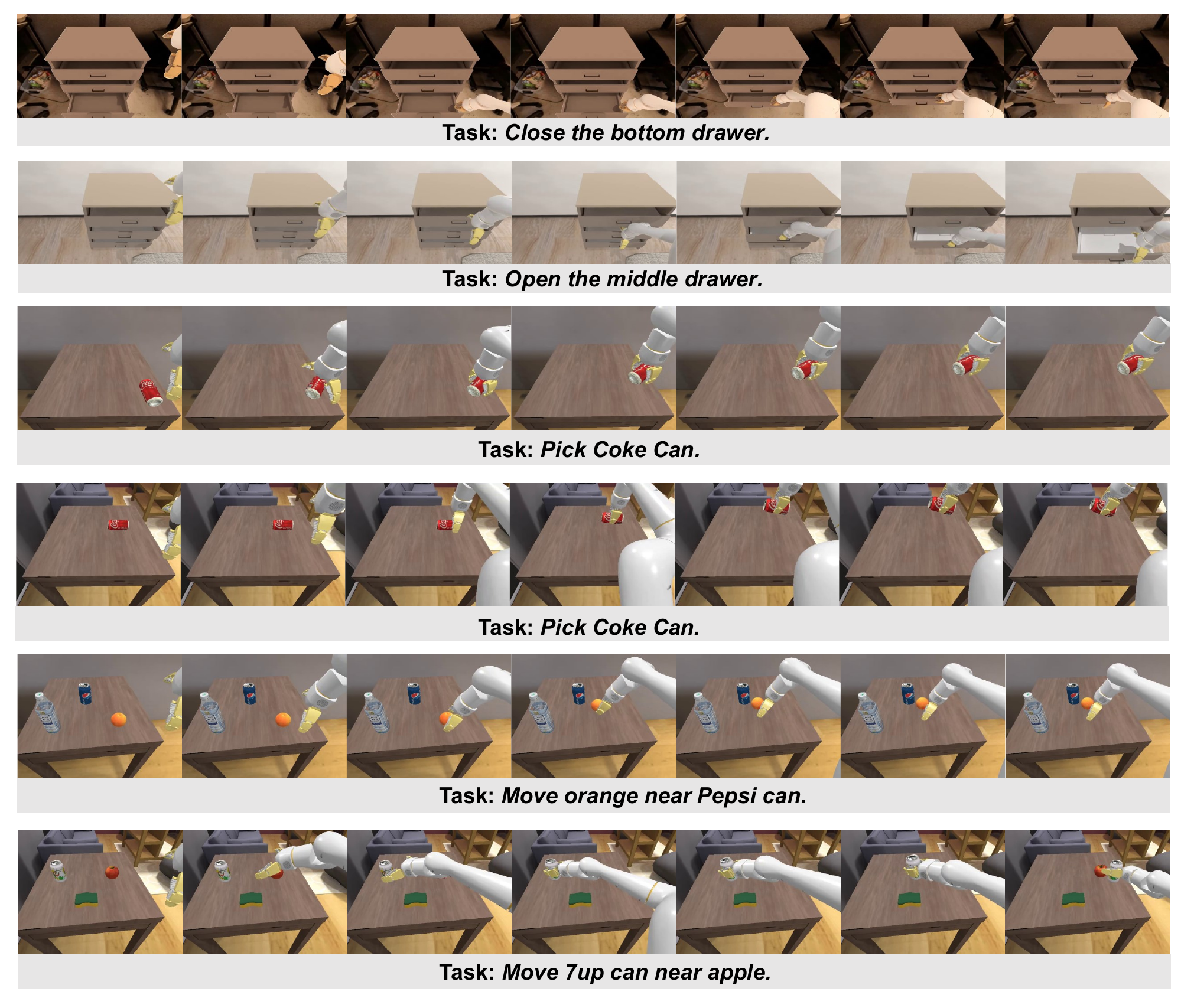}
  \caption{ \label{fig: vis_google} 
Visual examples of SimplerEnv Google robot tasks driven by {\modelname}.}
\end{figure*}

\begin{figure*}[t]
    \centering
    \includegraphics[width=0.95\textwidth]{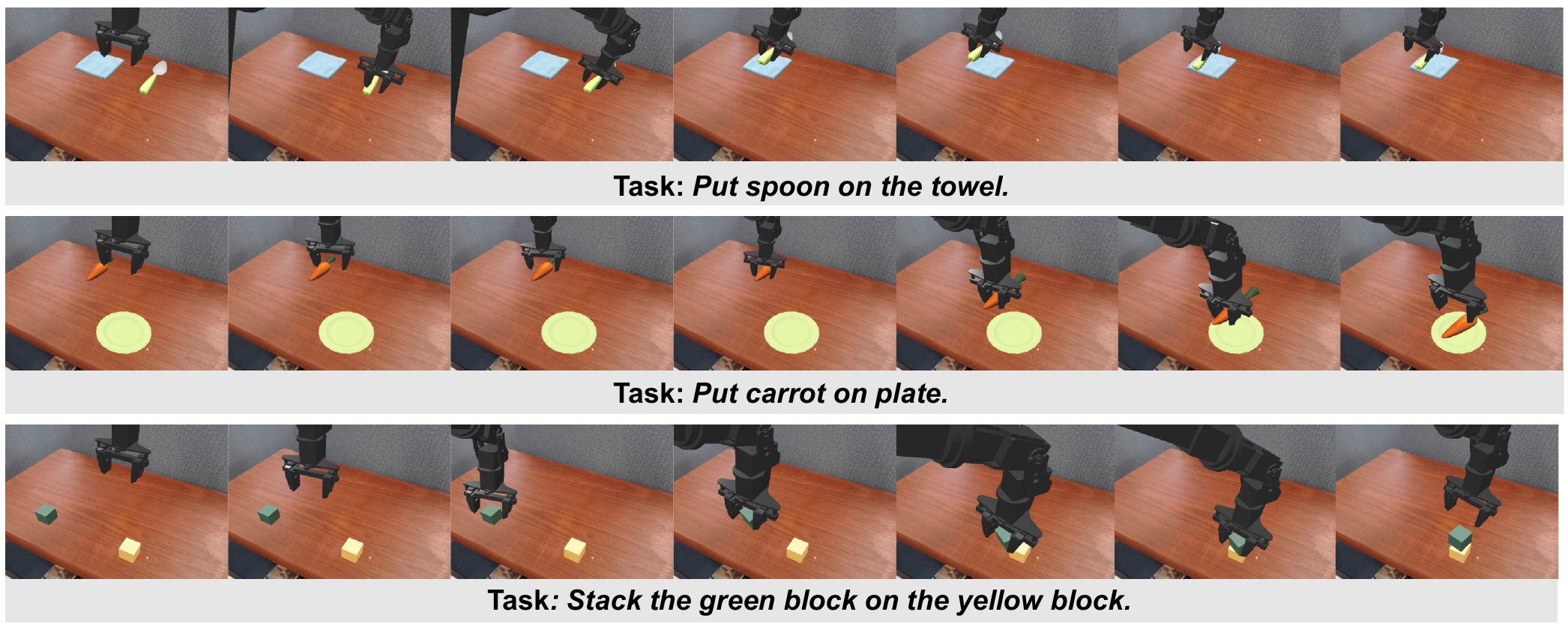}
  \caption{ \label{fig: vis_widowx} 
Visual examples of SimplerEnv WidowX robot tasks driven by {\modelname}.}
\end{figure*}

\begin{figure*}[t]
    \centering
    \includegraphics[width=0.95\textwidth]{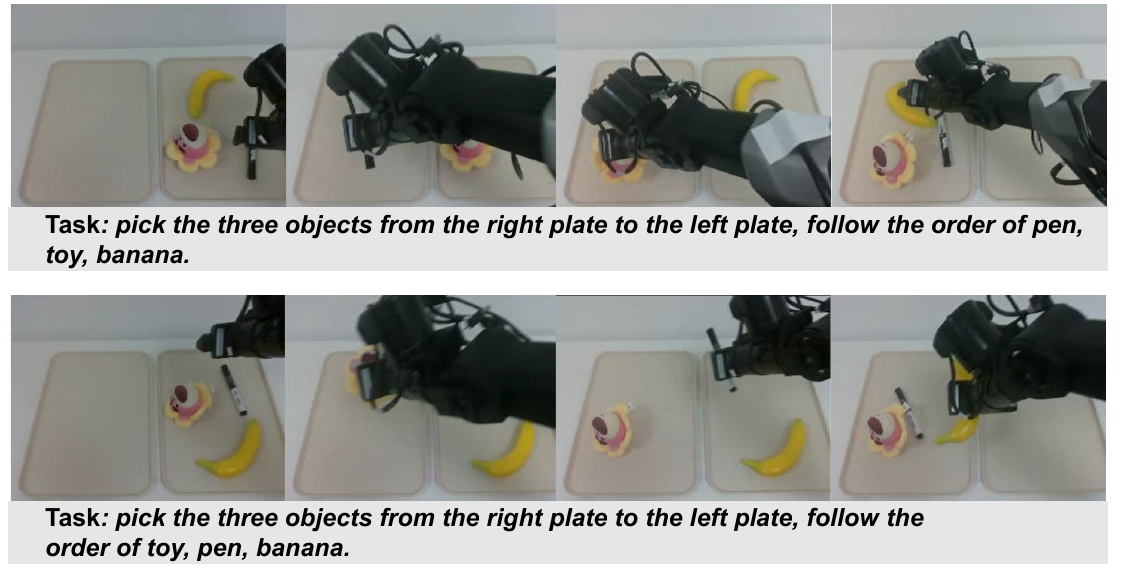}
  \caption{ \label{fig: real_success1} 
The successful cases of {\modelname} in real-world tasks.}
\end{figure*}

\begin{figure*}[t]
    \centering
    \includegraphics[width=0.95\textwidth]{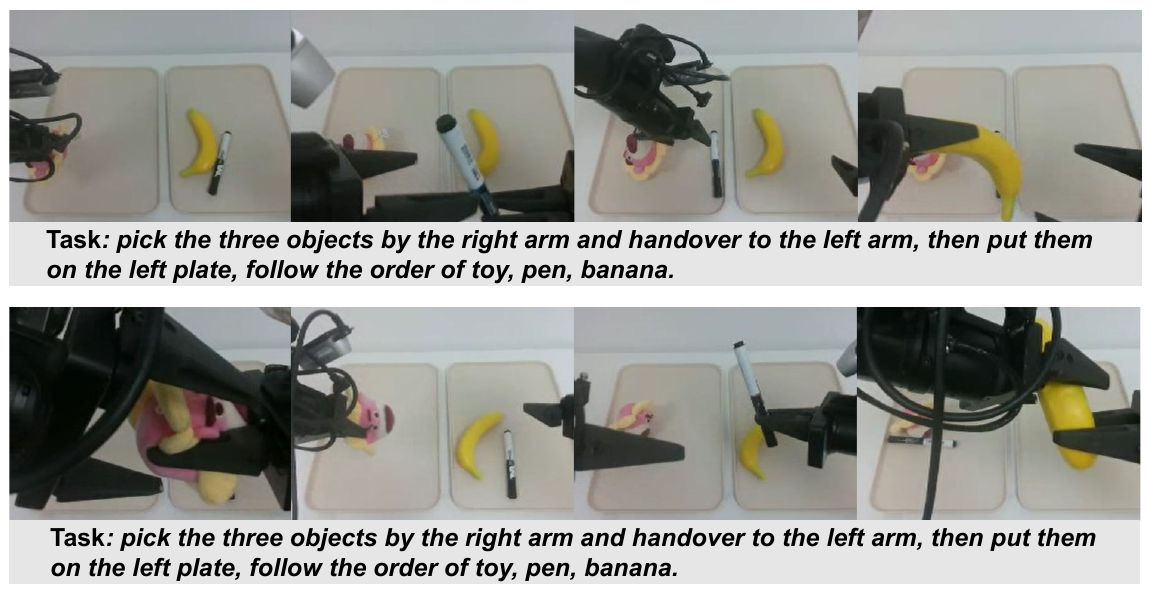}
  \caption{ \label{fig: real_success2} 
The successful cases of {\modelname} in real-world tasks.}
\end{figure*}

\paragraph{Simulation.}
Fig.~\ref{fig: vis_google} and Fig.~\ref{fig: vis_widowx} show successful execution examples of {\modelname} on SimplerEnv.
\paragraph{Real-Task.}
Fig.~\ref{fig: real_success1} and Fig.~\ref{fig: real_success2} show successful execution examples of {\modelname} on real-world tasks.
\subsection{Additional Failure Case Analysis
}
\begin{figure*}[t]
    \centering
    \includegraphics[width=0.95\textwidth]{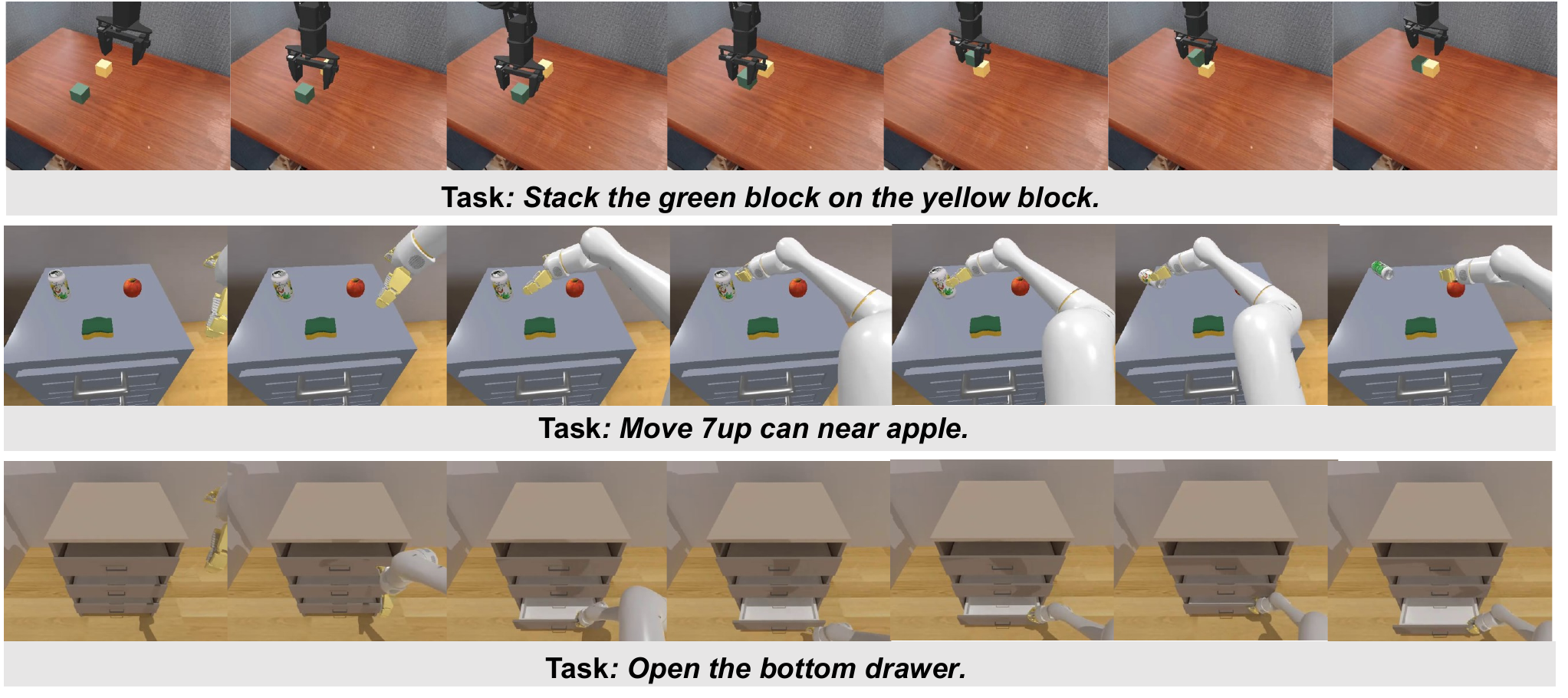}
  \caption{ \label{fig: failure_simpler} 
The failure cases of {\modelname} in SimplerEnv.}
\end{figure*}

\begin{figure*}[t]
    \centering
    \includegraphics[width=0.95\textwidth]{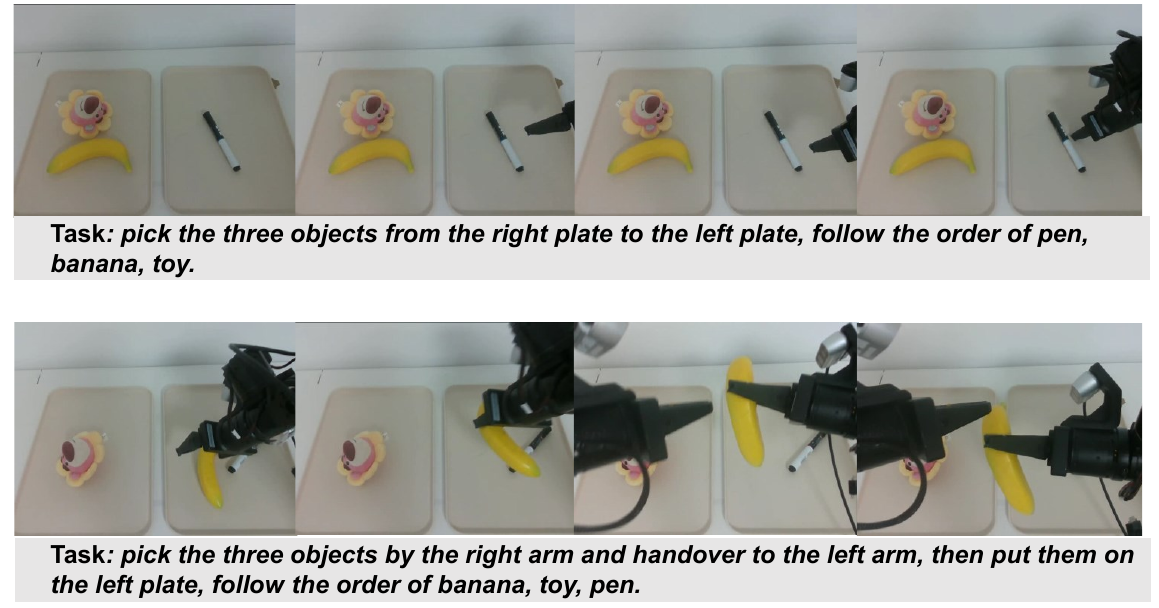}
  \caption{ \label{fig: real_fail} 
The failure cases of {\modelname} in real-world tasks.}
\end{figure*}

The failure cases of {\modelname}, illustrated in Fig.~\ref{fig: failure_simpler} and Fig.~\ref{fig: real_fail}, are primarily caused by insufficiently detailed action modeling, which can lead to imprecise or unstable execution, and the lack of historical context, which sometimes results in suboptimal or inconsistent decision-making. Additionally, a portion of failures arises from evaluation bias in the simulation environment, where tasks that are not perfectly completed are still marked as unsuccessful. These observations underscore the importance of fine-grained assessment, as provided by the proposed {\evalname}.
\clearpage


\begin{figure*}[!ht] 
\begin{AIbox}{{\evalname} for Reasoning VLAs}
{\color{black}\bf \large System Prompt} 
\vspace{1mm}
\\
\textbf{Character Introduction}  
\\
You are an \textbf{engineer proficient in evaluating robotic operations}.  
Your task is to perform a \textbf{fine-grained, rigorous evaluation} of a robot’s behavior.
\\
\textbf{Input Information}  
\\
You will receive the following four components:
\begin{enumerate}
  \item \textbf{Task Description} — natural language description of the robot’s assigned task.
  \item \textbf{Motion Trajectory} — a sequence of images representing the robot’s movements over time.
  \item \textbf{Reasoning Content} — the robot’s moment-by-moment reasoning corresponding to the trajectory.
\end{enumerate}
\textbf{Your Task}  
\\
Evaluate the robot’s performance along \textbf{four dimensions}, each scored on a \textbf{0–10} scale:

{
\centering
\small
\setlength{\tabcolsep}{8pt}
\renewcommand{\arraystretch}{1.2}
\begin{tabular}{|p{3.3cm}|p{11.5cm}|}
\hline
\textbf{Dimension} & \textbf{Description} \\ 
\hline
Reasoning Score & Measures the correctness, logical consistency, and usefulness of the reasoning in guiding the task. \\
\hline
Action Score & Measures the coherence, precision, and efficiency of the action sequence in achieving the task. \\
\hline
Intention Score & Evaluates whether the reasoning and actions constructively contribute toward solving the task. \\
\hline
Reason–Act Alignment Score & Measures the consistency between reasoning and corresponding actions, ensuring logical and behavioral alignment. \\
\hline
\end{tabular}
\par
}

\vspace{1mm}
\noindent
\textit{Scoring Guideline:} \quad 10 = excellent; \ 5 = acceptable; \ 0 = poor or missing evidence.

\textbf{Evaluation Requirements}
\begin{itemize}
  \item Be \textbf{objective}, \textbf{thorough}, and \textbf{specific} — do not overlook details or fabricate facts.
  \item If the task or data is ambiguous, \textbf{acknowledge uncertainty explicitly}.
  \item Mention any \textbf{critical failure modes} (e.g., collisions, unsafe motions, too slow).
  \item What you are seeing is the complete process of a robotic arm performing a task. \textbf{A task must be fully executed to be considered successful}; if opening a drawer, it must be fully opened, and if closing a drawer, it must be fully closed.
  \item Please provide a rigorous evaluation.
\end{itemize}
\textbf{Example}  
\textbf{Output Format}  
\\
Output must be \textbf{strictly formatted in JSON} and include your internal reasoning under \texttt{"Thought"}.

\begin{lstlisting}[style=prompt]
{
  "Thought": string, // Briefly describe your thought process and reasoning steps to evaluate the robot's performance.
  "Result": {
    "Action": int,
    "Intention": int,
    "Reasoning": int,
    "Alignment": int,
    "Success": bool [optional]
  }
}
\end{lstlisting}
\tcblower
{\color{deepblue}\bf \large User:} \\
Task: Pick Coke Can; Reasoning: [Reasoning Contents]; Trajectory:[Images]
\end{AIbox}
\vspace{-1em}
\caption{An example prompt of {\evalname} for reasoning VLAs.}
\label{fig:reasoning-eval-prompt}
\end{figure*}

\begin{figure*}[!ht] 
\begin{AIbox}{{\evalname} for Specialist VLAs}
{\color{black}\bf \large System Prompt} 
\vspace{1mm}
\\
\textbf{Character Introduction}  
\\
You are an \textbf{engineer proficient in evaluating robotic operations}.  
Your task is to perform a \textbf{fine-grained, rigorous evaluation} of a robot’s behavior.
\\
\textbf{Input Information}  
\\
You will receive the following four components:
\begin{enumerate}
  \item \textbf{Task Description} — natural language description of the robot’s assigned task.
  \item \textbf{Motion Trajectory} — a sequence of images representing the robot’s movements over time.
\end{enumerate}
\textbf{Your Task}  
\\
Evaluate the robot’s performance along \textbf{four dimensions}, each scored on a \textbf{0–10} scale:

{
\centering
\small
\setlength{\tabcolsep}{8pt}
\renewcommand{\arraystretch}{1.2}
\begin{tabular}{|p{3.3cm}|p{11.5cm}|}
\hline
\textbf{Dimension} & \textbf{Description} \\ 
\hline
Action Score & Measures the coherence, precision, and efficiency of the action sequence in achieving the task. \\
\hline
Intention Score & Evaluates whether the reasoning and actions constructively contribute toward solving the task. \\
\hline
\end{tabular}
\par
}

\vspace{1mm}
\noindent
\textit{Scoring Guideline:} \quad 10 = excellent; \ 5 = acceptable; \ 0 = poor or missing evidence.

\textbf{Evaluation Requirements}
\begin{itemize}
  \item Be \textbf{objective}, \textbf{thorough}, and \textbf{specific} — do not overlook details or fabricate facts.
  \item If the task or data is ambiguous, \textbf{acknowledge uncertainty explicitly}.
  \item Mention any \textbf{critical failure modes} (e.g., collisions, unsafe motions, too slow).
  \item What you are seeing is the complete process of a robotic arm performing a task. \textbf{A task must be fully executed to be considered successful}; if opening a drawer, it must be fully opened, and if closing a drawer, it must be fully closed.
  \item Please provide a rigorous evaluation.
\end{itemize}
\textbf{Examples} 
\textbf{Output Format}  
\\
Output must be \textbf{strictly formatted in JSON} and include your internal reasoning under \texttt{"Thought"}.

\begin{lstlisting}[style=prompt]
{
  "Thought": string, // Briefly describe your thought process and reasoning steps to evaluate the robot's performance.
  "Result": {
    "Action": int,
    "Intention": int,
    "Success": bool [optional]
  }
}
\end{lstlisting}

\tcblower
{\color{deepblue}\bf \large User:} \\
Task: Pick Coke Can; Trajectory:[Images]
\end{AIbox}
\vspace{-1em}
\caption{An example prompt of {\evalname} for specialist VLAs.}
\label{fig:specialist-eval-prompt}
\end{figure*}

%% file: main.bbl
\begin{thebibliography}{78}
\providecommand{\natexlab}[1]{#1}
\providecommand{\url}[1]{\texttt{#1}}
\expandafter\ifx\csname urlstyle\endcsname\relax
  \providecommand{\doi}[1]{doi: #1}\else
  \providecommand{\doi}{doi: \begingroup \urlstyle{rm}\Url}\fi

\bibitem[Arnab et~al.(2025)Arnab, Iscen, Caron, Fathi, and
  Schmid]{arnab2025temporal}
Anurag Arnab, Ahmet Iscen, Mathilde Caron, Alireza Fathi, and Cordelia Schmid.
\newblock Temporal chain of thought: Long-video understanding by thinking in
  frames.
\newblock \emph{arXiv preprint arXiv:2507.02001}, 2025.

\bibitem[Black et~al.(2024)Black, Brown, Driess, Esmail, Equi, Finn, Fusai,
  Groom, Hausman, Ichter, et~al.]{pi0}
Kevin Black, Noah Brown, Danny Driess, Adnan Esmail, Michael Equi, Chelsea
  Finn, Niccolo Fusai, Lachy Groom, Karol Hausman, Brian Ichter, et~al.
\newblock $\pi_0$: A vision-language-action flow model for general robot
  control.
\newblock \emph{arXiv preprint arXiv:2410.24164}, 2024.

\bibitem[Chen et~al.(2024{\natexlab{a}})Chen, Chen, Zhang, Wang, Liu, Zhou,
  Zhang, Wan, Zhou, and Sun]{mllmasajudge}
Dongping Chen, Ruoxi Chen, Shilin Zhang, Yaochen Wang, Yinuo Liu, Huichi Zhou,
  Qihui Zhang, Yao Wan, Pan Zhou, and Lichao Sun.
\newblock {MLLM}-as-a-judge: Assessing multimodal {LLM}-as-a-judge with
  vision-language benchmark.
\newblock In \emph{Forty-first International Conference on Machine Learning},
  2024{\natexlab{a}}.

\bibitem[Chen et~al.(2025{\natexlab{a}})Chen, Liu, Gu, Liu, Zhang, Li, He, Guo,
  Fu, Zhang, et~al.]{chen2025fast}
Hao Chen, Jiaming Liu, Chenyang Gu, Zhuoyang Liu, Renrui Zhang, Xiaoqi Li, Xiao
  He, Yandong Guo, Chi-Wing Fu, Shanghang Zhang, et~al.
\newblock Fast-in-slow: A dual-system foundation model unifying fast
  manipulation within slow reasoning.
\newblock \emph{arXiv preprint arXiv:2506.01953}, 2025{\natexlab{a}}.

\bibitem[Chen et~al.(2024{\natexlab{b}})Chen, Li, Dong, Zhang, Zang, Chen,
  Duan, Wang, Qiao, Lin, et~al.]{MMSTAR}
Lin Chen, Jinsong Li, Xiaoyi Dong, Pan Zhang, Yuhang Zang, Zehui Chen, Haodong
  Duan, Jiaqi Wang, Yu Qiao, Dahua Lin, et~al.
\newblock Are we on the right way for evaluating large vision-language models?
\newblock \emph{Advances in Neural Information Processing Systems},
  37:\penalty0 27056--27087, 2024{\natexlab{b}}.

\bibitem[Chen et~al.(2025{\natexlab{b}})Chen, Mao, Gu, and Shou]{chen2025edit}
Lan Chen, Qi Mao, Yuchao Gu, and Mike~Zheng Shou.
\newblock Edit transfer: Learning image editing via vision in-context
  relations.
\newblock \emph{arXiv preprint arXiv:2503.13327}, 2025{\natexlab{b}}.

\bibitem[Chen et~al.(2025{\natexlab{c}})Chen, Belkhale, Mirchandani, Mees,
  Driess, Pertsch, and Levine]{ecot-lite}
William Chen, Suneel Belkhale, Suvir Mirchandani, Oier Mees, Danny Driess, Karl
  Pertsch, and Sergey Levine.
\newblock Training strategies for efficient embodied reasoning.
\newblock \emph{arXiv preprint arXiv:2505.08243}, 2025{\natexlab{c}}.

\bibitem[Chen et~al.(2024{\natexlab{c}})Chen, Xu, Liang, He, Pang, Yu, Song,
  Liu, Zhou, Zhang, et~al.]{chen2024not}
Xingyu Chen, Jiahao Xu, Tian Liang, Zhiwei He, Jianhui Pang, Dian Yu, Linfeng
  Song, Qiuzhi Liu, Mengfei Zhou, Zhuosheng Zhang, et~al.
\newblock Do not think that much for 2+ 3=? on the overthinking of o1-like
  llms.
\newblock \emph{arXiv preprint arXiv:2412.21187}, 2024{\natexlab{c}}.

\bibitem[Duan et~al.(2025)Duan, Zhang, Geng, Liu, Boedecker, and Lu]{fast-ecot}
Zhekai Duan, Yuan Zhang, Shikai Geng, Gaowen Liu, Joschka Boedecker, and
  Chris~Xiaoxuan Lu.
\newblock Fast ecot: Efficient embodied chain-of-thought via thoughts reuse.
\newblock \emph{arXiv preprint arXiv:2506.07639}, 2025.

\bibitem[Farrow and Abernethy(2003)]{farrow2003expertise}
Damian Farrow and Bruce Abernethy.
\newblock Do expertise and the degree of perception—action coupling affect
  natural anticipatory performance?
\newblock \emph{Perception}, 32\penalty0 (9):\penalty0 1127--1139, 2003.

\bibitem[Gu et~al.(2024)Gu, Jiang, Shi, Tan, Zhai, Xu, Li, Shen, Ma, Liu,
  et~al.]{gu2024survey}
Jiawei Gu, Xuhui Jiang, Zhichao Shi, Hexiang Tan, Xuehao Zhai, Chengjin Xu, Wei
  Li, Yinghan Shen, Shengjie Ma, Honghao Liu, et~al.
\newblock A survey on llm-as-a-judge.
\newblock \emph{arXiv preprint arXiv:2411.15594}, 2024.

\bibitem[Huang et~al.(2025{\natexlab{a}})Huang, Wu, Chen, Wang, and
  Yang]{thinkact}
Chi-Pin Huang, Yueh-Hua Wu, Min-Hung Chen, Yu-Chiang~Frank Wang, and Fu-En
  Yang.
\newblock Thinkact: Vision-language-action reasoning via reinforced visual
  latent planning.
\newblock \emph{arXiv preprint arXiv:2507.16815}, 2025{\natexlab{a}}.

\bibitem[Huang et~al.(2025{\natexlab{b}})Huang, Fang, Chen, Yuan, Ye, Zeng,
  Chen, Mao, and Zhao]{critictool}
Shiting Huang, Zhen Fang, Zehui Chen, Siyu Yuan, Junjie Ye, Yu Zeng, Lin Chen,
  Qi Mao, and Feng Zhao.
\newblock Critictool: Evaluating self-critique capabilities of large language
  models in tool-calling error scenarios.
\newblock \emph{arXiv preprint arXiv:2506.13977}, 2025{\natexlab{b}}.

\bibitem[Huang et~al.(2025{\natexlab{c}})Huang, Chen, Xie, Cao, Tang, Shen,
  Yin, Hu, Wang, Tang, et~al.]{huang2025interleaving}
Wenxuan Huang, Shuang Chen, Zheyong Xie, Shaosheng Cao, Shixiang Tang, Yufan
  Shen, Qingyu Yin, Wenbo Hu, Xiaoman Wang, Yuntian Tang, et~al.
\newblock Interleaving reasoning for better text-to-image generation.
\newblock \emph{arXiv preprint arXiv:2509.06945}, 2025{\natexlab{c}}.

\bibitem[Huang et~al.(2024)Huang, He, Yu, Zhang, Si, Jiang, Zhang, Wu, Jin,
  Chanpaisit, et~al.]{huang2024vbench}
Ziqi Huang, Yinan He, Jiashuo Yu, Fan Zhang, Chenyang Si, Yuming Jiang, Yuanhan
  Zhang, Tianxing Wu, Qingyang Jin, Nattapol Chanpaisit, et~al.
\newblock Vbench: Comprehensive benchmark suite for video generative models.
\newblock In \emph{Proceedings of the IEEE/CVF Conference on Computer Vision
  and Pattern Recognition}, pages 21807--21818, 2024.

\bibitem[Huang et~al.(2025{\natexlab{d}})Huang, Liu, Lin, Zhu, Zhao, Du, Li,
  Jia, Zhong, Chen, et~al.]{notvla}
Zheng Huang, Mingyu Liu, Xiaoyi Lin, Muzhi Zhu, Canyu Zhao, Zongze Du, Xiaoman
  Li, Yiduo Jia, Hao Zhong, Hao Chen, et~al.
\newblock Notvla: Narrowing of dense action trajectories for generalizable
  robot manipulation.
\newblock \emph{arXiv preprint arXiv:2510.03895}, 2025{\natexlab{d}}.

\bibitem[Hurst et~al.(2024)Hurst, Lerer, Goucher, Perelman, Ramesh, Clark,
  Ostrow, Welihinda, Hayes, Radford, et~al.]{gpt4o}
Aaron Hurst, Adam Lerer, Adam~P Goucher, Adam Perelman, Aditya Ramesh, Aidan
  Clark, AJ Ostrow, Akila Welihinda, Alan Hayes, Alec Radford, et~al.
\newblock Gpt-4o system card.
\newblock \emph{arXiv preprint arXiv:2410.21276}, 2024.

\bibitem[Jia et~al.(2025)Jia, Qi, Zhang, Zhang, Yu, He, Wang, and
  Yi]{jia2025omnispatial}
Mengdi Jia, Zekun Qi, Shaochen Zhang, Wenyao Zhang, Xinqiang Yu, Jiawei He, He
  Wang, and Li Yi.
\newblock Omnispatial: Towards comprehensive spatial reasoning benchmark for
  vision language models.
\newblock \emph{arXiv preprint arXiv:2506.03135}, 2025.

\bibitem[Jiang et~al.(2025)Jiang, Zhang, Guo, Li, Qi, Chen, Wang, Jin, Guo,
  Yan, et~al.]{jiang2025mme}
Dongzhi Jiang, Renrui Zhang, Ziyu Guo, Yanwei Li, Yu Qi, Xinyan Chen, Liuhui
  Wang, Jianhan Jin, Claire Guo, Shen Yan, et~al.
\newblock Mme-cot: Benchmarking chain-of-thought in large multimodal models for
  reasoning quality, robustness, and efficiency.
\newblock \emph{arXiv preprint arXiv:2502.09621}, 2025.

\bibitem[Kaplan et~al.(2020)Kaplan, McCandlish, Henighan, Brown, Chess, Child,
  Gray, Radford, Wu, and Amodei]{scalinglaw}
Jared Kaplan, Sam McCandlish, Tom Henighan, Tom~B Brown, Benjamin Chess, Rewon
  Child, Scott Gray, Alec Radford, Jeffrey Wu, and Dario Amodei.
\newblock Scaling laws for neural language models.
\newblock \emph{arXiv preprint arXiv:2001.08361}, 2020.

\bibitem[Karamcheti et~al.(2024)Karamcheti, Nair, Balakrishna, Liang, Kollar,
  and Sadigh]{karamcheti2024prismatic}
Siddharth Karamcheti, Suraj Nair, Ashwin Balakrishna, Percy Liang, Thomas
  Kollar, and Dorsa Sadigh.
\newblock Prismatic vlms: Investigating the design space of
  visually-conditioned language models.
\newblock In \emph{Forty-first International Conference on Machine Learning},
  2024.

\bibitem[Khazatsky et~al.(2025)Khazatsky, Pertsch, Nair, Balakrishna, Dasari,
  Karamcheti, Nasiriany, Srirama, Chen, Ellis, Fagan, Hejna, Itkina, Lepert,
  Ma, Miller, Wu, Belkhale, Dass, Ha, Jain, Lee, Lee, Memmel, Park,
  Radosavovic, Wang, Zhan, Black, Chi, Hatch, Lin, Lu, Mercat, Rehman, Sanketi,
  Sharma, Simpson, Vuong, Walke, Wulfe, Xiao, Yang, Yavary, Zhao, Agia, Baijal,
  Castro, Chen, Chen, Chung, Drake, Foster, Gao, Guizilini, Herrera, Heo, Hsu,
  Hu, Irshad, Jackson, Le, Li, Lin, Lin, Ma, Maddukuri, Mirchandani, Morton,
  Nguyen, O'Neill, Scalise, Seale, Son, Tian, Tran, Wang, Wu, Xie, Yang, Yin,
  Zhang, Bastani, Berseth, Bohg, Goldberg, Gupta, Gupta, Jayaraman, Lim, Malik,
  Martín-Martín, Ramamoorthy, Sadigh, Song, Wu, Yip, Zhu, Kollar, Levine, and
  Finn]{droid}
Alexander Khazatsky, Karl Pertsch, Suraj Nair, Ashwin Balakrishna, Sudeep
  Dasari, Siddharth Karamcheti, Soroush Nasiriany, Mohan~Kumar Srirama,
  Lawrence~Yunliang Chen, Kirsty Ellis, Peter~David Fagan, Joey Hejna, Masha
  Itkina, Marion Lepert, Yecheng~Jason Ma, Patrick~Tree Miller, Jimmy Wu,
  Suneel Belkhale, Shivin Dass, Huy Ha, Arhan Jain, Abraham Lee, Youngwoon Lee,
  Marius Memmel, Sungjae Park, Ilija Radosavovic, Kaiyuan Wang, Albert Zhan,
  Kevin Black, Cheng Chi, Kyle~Beltran Hatch, Shan Lin, Jingpei Lu, Jean
  Mercat, Abdul Rehman, Pannag~R Sanketi, Archit Sharma, Cody Simpson, Quan
  Vuong, Homer~Rich Walke, Blake Wulfe, Ted Xiao, Jonathan~Heewon Yang, Arefeh
  Yavary, Tony~Z. Zhao, Christopher Agia, Rohan Baijal, Mateo~Guaman Castro,
  Daphne Chen, Qiuyu Chen, Trinity Chung, Jaimyn Drake, Ethan~Paul Foster,
  Jensen Gao, Vitor Guizilini, David~Antonio Herrera, Minho Heo, Kyle Hsu,
  Jiaheng Hu, Muhammad~Zubair Irshad, Donovon Jackson, Charlotte Le, Yunshuang
  Li, Kevin Lin, Roy Lin, Zehan Ma, Abhiram Maddukuri, Suvir Mirchandani,
  Daniel Morton, Tony Nguyen, Abigail O'Neill, Rosario Scalise, Derick Seale,
  Victor Son, Stephen Tian, Emi Tran, Andrew~E. Wang, Yilin Wu, Annie Xie,
  Jingyun Yang, Patrick Yin, Yunchu Zhang, Osbert Bastani, Glen Berseth,
  Jeannette Bohg, Ken Goldberg, Abhinav Gupta, Abhishek Gupta, Dinesh
  Jayaraman, Joseph~J Lim, Jitendra Malik, Roberto Martín-Martín, Subramanian
  Ramamoorthy, Dorsa Sadigh, Shuran Song, Jiajun Wu, Michael~C. Yip, Yuke Zhu,
  Thomas Kollar, Sergey Levine, and Chelsea Finn.
\newblock Droid: A large-scale in-the-wild robot manipulation dataset, 2025.

\bibitem[Kim et~al.(2024)Kim, Pertsch, Karamcheti, Xiao, Balakrishna, Nair,
  Rafailov, Foster, Lam, Sanketi, et~al.]{openvla}
Moo~Jin Kim, Karl Pertsch, Siddharth Karamcheti, Ted Xiao, Ashwin Balakrishna,
  Suraj Nair, Rafael Rafailov, Ethan Foster, Grace Lam, Pannag Sanketi, et~al.
\newblock Openvla: An open-source vision-language-action model.
\newblock \emph{arXiv preprint arXiv:2406.09246}, 2024.

\bibitem[Lee et~al.(2025)Lee, Duan, Fang, Deng, Liu, Li, Fang, Zhang, Wang,
  Lee, et~al.]{molmoact}
Jason Lee, Jiafei Duan, Haoquan Fang, Yuquan Deng, Shuo Liu, Boyang Li, Bohan
  Fang, Jieyu Zhang, Yi~Ru Wang, Sangho Lee, et~al.
\newblock Molmoact: Action reasoning models that can reason in space.
\newblock \emph{arXiv preprint arXiv:2508.07917}, 2025.

\bibitem[Li et~al.(2025{\natexlab{a}})Li, Yang, Chen, Tian, Yang, Chen, Wang,
  Wang, Zhao, Lin, et~al.]{li2025cronusvla}
Hao Li, Shuai Yang, Yilun Chen, Yang Tian, Xiaoda Yang, Xinyi Chen, Hanqing
  Wang, Tai Wang, Feng Zhao, Dahua Lin, et~al.
\newblock Cronusvla: Transferring latent motion across time for multi-frame
  prediction in manipulation.
\newblock \emph{arXiv preprint arXiv:2506.19816}, 2025{\natexlab{a}}.

\bibitem[Li et~al.(2023)Li, Wei, Han, and Fan]{Li_2023_ICCV}
Jiapeng Li, Ping Wei, Wenjuan Han, and Lifeng Fan.
\newblock Intentqa: Context-aware video intent reasoning.
\newblock In \emph{Proceedings of the IEEE/CVF International Conference on
  Computer Vision (ICCV)}, pages 11963--11974, 2023.

\bibitem[Li et~al.(2024{\natexlab{a}})Li, Liang, Wang, Luo, Chen, Liao, Wei,
  Deng, Xu, Zhang, et~al.]{cogact}
Qixiu Li, Yaobo Liang, Zeyu Wang, Lin Luo, Xi Chen, Mozheng Liao, Fangyun Wei,
  Yu Deng, Sicheng Xu, Yizhong Zhang, et~al.
\newblock Cogact: A foundational vision-language-action model for synergizing
  cognition and action in robotic manipulation.
\newblock \emph{arXiv preprint arXiv:2411.19650}, 2024{\natexlab{a}}.

\bibitem[Li et~al.(2024{\natexlab{b}})Li, Hsu, Gu, Mees, Pertsch, Walke, Fu,
  Lunawat, Sieh, Kirmani, Levine, Wu, Finn, Su, Vuong, and Xiao]{simplerenv}
Xuanlin Li, Kyle Hsu, Jiayuan Gu, Oier Mees, Karl Pertsch, Homer~Rich Walke,
  Chuyuan Fu, Ishikaa Lunawat, Isabel Sieh, Sean Kirmani, Sergey Levine, Jiajun
  Wu, Chelsea Finn, Hao Su, Quan Vuong, and Ted Xiao.
\newblock Evaluating real-world robot manipulation policies in simulation.
\newblock In \emph{8th Annual Conference on Robot Learning},
  2024{\natexlab{b}}.

\bibitem[Li et~al.(2025{\natexlab{b}})Li, Xu, Zhang, Liu, Shen, Ponomarenko,
  Xu, Heng, Huang, Zhang, and Dong]{Li_2025_CVPR}
Xiaoqi Li, Jingyun Xu, Mingxu Zhang, Jiaming Liu, Yan Shen, Iaroslav
  Ponomarenko, Jiahui Xu, Liang Heng, Siyuan Huang, Shanghang Zhang, and Hao
  Dong.
\newblock Object-centric prompt-driven vision-language-action model for robotic
  manipulation.
\newblock In \emph{Proceedings of the IEEE/CVF Conference on Computer Vision
  and Pattern Recognition (CVPR)}, pages 27638--27648, 2025{\natexlab{b}}.

\bibitem[Li(2025)]{li2025advancing}
Yanshu Li.
\newblock Advancing multimodal in-context learning in large vision-language
  models with task-aware demonstrations.
\newblock \emph{arXiv preprint arXiv:2503.04839}, 2025.

\bibitem[Li et~al.(2025{\natexlab{c}})Li, Yu, Huang, Liu, Liang, Liu, Che, Yu,
  Boyd-Graber, Mi, et~al.]{li2025self}
Zongxia Li, Wenhao Yu, Chengsong Huang, Rui Liu, Zhenwen Liang, Fuxiao Liu,
  Jingxi Che, Dian Yu, Jordan Boyd-Graber, Haitao Mi, et~al.
\newblock Self-rewarding vision-language model via reasoning decomposition.
\newblock \emph{arXiv preprint arXiv:2508.19652}, 2025{\natexlab{c}}.

\bibitem[Lin et~al.(2025)Lin, Nai, Hu, You, Zhao, and Gao]{onetwovla}
Fanqi Lin, Ruiqian Nai, Yingdong Hu, Jiacheng You, Junming Zhao, and Yang Gao.
\newblock Onetwovla: A unified vision-language-action model with adaptive
  reasoning.
\newblock \emph{arXiv preprint arXiv:2505.11917}, 2025.

\bibitem[Liu et~al.(2023)Liu, Zhu, Gao, Feng, Liu, Zhu, and Stone]{libero}
Bo Liu, Yifeng Zhu, Chongkai Gao, Yihao Feng, Qiang Liu, Yuke Zhu, and Peter
  Stone.
\newblock Libero: Benchmarking knowledge transfer for lifelong robot learning.
\newblock \emph{Advances in Neural Information Processing Systems},
  36:\penalty0 44776--44791, 2023.

\bibitem[Liu et~al.(2025{\natexlab{a}})Liu, Li, Li, Liu, Wang, Liu, Kang, Ma,
  Kong, and Zhang]{robovlms}
Huaping Liu, Xinghang Li, Peiyan Li, Minghuan Liu, Dong Wang, Jirong Liu,
  Bingyi Kang, Xiao Ma, Tao Kong, and Hanbo Zhang.
\newblock Towards generalist robot policies: What matters in building
  vision-language-action models.
\newblock 2025{\natexlab{a}}.

\bibitem[Liu et~al.(2025{\natexlab{b}})Liu, Chen, An, Liu, Zhang, Gu, Li, Guo,
  Chen, Liu, et~al.]{hybridvla}
Jiaming Liu, Hao Chen, Pengju An, Zhuoyang Liu, Renrui Zhang, Chenyang Gu,
  Xiaoqi Li, Ziyu Guo, Sixiang Chen, Mengzhen Liu, et~al.
\newblock Hybridvla: Collaborative diffusion and autoregression in a unified
  vision-language-action model.
\newblock \emph{arXiv preprint arXiv:2503.10631}, 2025{\natexlab{b}}.

\bibitem[Liu et~al.(2024{\natexlab{a}})Liu, Duan, Zhang, Li, Zhang, Zhao, Yuan,
  Wang, He, Liu, et~al.]{MMB}
Yuan Liu, Haodong Duan, Yuanhan Zhang, Bo Li, Songyang Zhang, Wangbo Zhao, Yike
  Yuan, Jiaqi Wang, Conghui He, Ziwei Liu, et~al.
\newblock Mmbench: Is your multi-modal model an all-around player?
\newblock In \emph{European conference on computer vision}, pages 216--233.
  Springer, 2024{\natexlab{a}}.

\bibitem[Liu et~al.(2024{\natexlab{b}})Liu, Li, Huang, Yang, Yu, Li, Yin, Liu,
  Jin, and Bai]{Ocrbench}
Yuliang Liu, Zhang Li, Mingxin Huang, Biao Yang, Wenwen Yu, Chunyuan Li,
  Xu-Cheng Yin, Cheng-Lin Liu, Lianwen Jin, and Xiang Bai.
\newblock Ocrbench: on the hidden mystery of ocr in large multimodal models.
\newblock \emph{Science China Information Sciences}, 67\penalty0 (12):\penalty0
  220102, 2024{\natexlab{b}}.

\bibitem[Liu et~al.(2025{\natexlab{c}})Liu, Gu, Zheng, Xue, and
  Fu]{liu2025trivla}
Zhenyang Liu, Yongchong Gu, Sixiao Zheng, Xiangyang Xue, and Yanwei Fu.
\newblock Trivla: A unified triple-system-based unified vision-language-action
  model for general robot control.
\newblock \emph{arXiv preprint arXiv:2507.01424}, 2025{\natexlab{c}}.

\bibitem[Liu et~al.(2025{\natexlab{d}})Liu, Liu, Xu, Han, Gu, Chen, Zhou,
  Zhang, Hsieh, Wu, et~al.]{mla}
Zhuoyang Liu, Jiaming Liu, Jiadong Xu, Nuowei Han, Chenyang Gu, Hao Chen,
  Kaichen Zhou, Renrui Zhang, Kai~Chin Hsieh, Kun Wu, et~al.
\newblock Mla: A multisensory language-action model for multimodal
  understanding and forecasting in robotic manipulation.
\newblock \emph{arXiv preprint arXiv:2509.26642}, 2025{\natexlab{d}}.

\bibitem[Mathew et~al.(2021)Mathew, Karatzas, and Jawahar]{DOCVQA}
Minesh Mathew, Dimosthenis Karatzas, and CV Jawahar.
\newblock Docvqa: A dataset for vqa on document images.
\newblock In \emph{Proceedings of the IEEE/CVF winter conference on
  applications of computer vision}, pages 2200--2209, 2021.

\bibitem[Mathew et~al.(2022)Mathew, Bagal, Tito, Karatzas, Valveny, and
  Jawahar]{infovqa}
Minesh Mathew, Viraj Bagal, Rub{\`e}n Tito, Dimosthenis Karatzas, Ernest
  Valveny, and CV Jawahar.
\newblock Infographicvqa.
\newblock In \emph{Proceedings of the IEEE/CVF Winter Conference on
  Applications of Computer Vision}, pages 1697--1706, 2022.

\bibitem[Mees et~al.(2022)Mees, Hermann, Rosete-Beas, and Burgard]{calvin}
Oier Mees, Lukas Hermann, Erick Rosete-Beas, and Wolfram Burgard.
\newblock Calvin: A benchmark for language-conditioned policy learning for
  long-horizon robot manipulation tasks.
\newblock \emph{IEEE Robotics and Automation Letters}, 7\penalty0 (3):\penalty0
  7327--7334, 2022.

\bibitem[Nguyen et~al.(2024)Nguyen, Luo, Shiri, Phung, Li, Vu, and
  Haffari]{nguyen2024direct}
Minh-Vuong Nguyen, Linhao Luo, Fatemeh Shiri, Dinh Phung, Yuan-Fang Li,
  Thuy-Trang Vu, and Gholamreza Haffari.
\newblock Direct evaluation of chain-of-thought in multi-hop reasoning with
  knowledge graphs.
\newblock \emph{arXiv preprint arXiv:2402.11199}, 2024.

\bibitem[O’Neill et~al.(2024)O’Neill, Rehman, Maddukuri, Gupta, Padalkar,
  Lee, Pooley, Gupta, Mandlekar, Jain, et~al.]{oxe-dataset}
Abby O’Neill, Abdul Rehman, Abhiram Maddukuri, Abhishek Gupta, Abhishek
  Padalkar, Abraham Lee, Acorn Pooley, Agrim Gupta, Ajay Mandlekar, Ajinkya
  Jain, et~al.
\newblock Open x-embodiment: Robotic learning datasets and rt-x models: Open
  x-embodiment collaboration 0.
\newblock In \emph{2024 IEEE International Conference on Robotics and
  Automation (ICRA)}, pages 6892--6903. IEEE, 2024.

\bibitem[Park et~al.(2020)Park, Efros, Zhang, and Zhu]{clip}
Taesung Park, Alexei~A Efros, Richard Zhang, and Jun-Yan Zhu.
\newblock Contrastive learning for unpaired image-to-image translation.
\newblock In \emph{European conference on computer vision}, pages 319--345.
  Springer, 2020.

\bibitem[Qi et~al.(2025)Qi, Zhao, Zeng, Bao, Huang, Chen, Chen, Zhao, Qi, and
  Zhao]{qi2025vcr}
Yukun Qi, Yiming Zhao, Yu Zeng, Xikun Bao, Wenxuan Huang, Lin Chen, Zehui Chen,
  Jie Zhao, Zhongang Qi, and Feng Zhao.
\newblock Vcr-bench: A comprehensive evaluation framework for video
  chain-of-thought reasoning.
\newblock \emph{arXiv preprint arXiv:2504.07956}, 2025.

\bibitem[Qin et~al.(2023)Qin, Liang, Ye, Zhu, Yan, Lu, Lin, Cong, Tang, Qian,
  et~al.]{toolbench}
Yujia Qin, Shihao Liang, Yining Ye, Kunlun Zhu, Lan Yan, Yaxi Lu, Yankai Lin,
  Xin Cong, Xiangru Tang, Bill Qian, et~al.
\newblock Toolllm: Facilitating large language models to master 16000+
  real-world apis.
\newblock \emph{arXiv preprint arXiv:2307.16789}, 2023.

\bibitem[Qu et~al.(2025)Qu, Song, Chen, Yao, Ye, Ding, Wang, Gu, Zhao, Wang,
  et~al.]{spatialvla}
Delin Qu, Haoming Song, Qizhi Chen, Yuanqi Yao, Xinyi Ye, Yan Ding, Zhigang
  Wang, JiaYuan Gu, Bin Zhao, Dong Wang, et~al.
\newblock Spatialvla: Exploring spatial representations for
  visual-language-action model.
\newblock \emph{arXiv preprint arXiv:2501.15830}, 2025.

\bibitem[Shukor et~al.(2025)Shukor, Aubakirova, Capuano, Kooijmans, Palma,
  Zouitine, Aractingi, Pascal, Russi, Marafioti, et~al.]{shukor2025smolvla}
Mustafa Shukor, Dana Aubakirova, Francesco Capuano, Pepijn Kooijmans, Steven
  Palma, Adil Zouitine, Michel Aractingi, Caroline Pascal, Martino Russi,
  Andres Marafioti, et~al.
\newblock Smolvla: A vision-language-action model for affordable and efficient
  robotics.
\newblock \emph{arXiv preprint arXiv:2506.01844}, 2025.

\bibitem[Singh et~al.(2019)Singh, Natarjan, Shah, Jiang, Chen, Parikh, and
  Rohrbach]{TextVQA}
Amanpreet Singh, Vivek Natarjan, Meet Shah, Yu Jiang, Xinlei Chen, Devi Parikh,
  and Marcus Rohrbach.
\newblock Towards vqa models that can read.
\newblock In \emph{Proceedings of the IEEE Conference on Computer Vision and
  Pattern Recognition}, pages 8317--8326, 2019.

\bibitem[Stephan et~al.(2024)Stephan, Zhu, A{\ss}enmacher, Shen, and
  Roth]{stephan2024calculation}
Andreas Stephan, Dawei Zhu, Matthias A{\ss}enmacher, Xiaoyu Shen, and Benjamin
  Roth.
\newblock From calculation to adjudication: Examining llm judges on
  mathematical reasoning tasks.
\newblock \emph{arXiv preprint arXiv:2409.04168}, 2024.

\bibitem[Sun et~al.(2024)Sun, Hong, Pala, Toh, Tan, Ghosal, Poria,
  et~al.]{emma-x}
Qi Sun, Pengfei Hong, Tej~Deep Pala, Vernon Toh, U Tan, Deepanway Ghosal,
  Soujanya Poria, et~al.
\newblock Emma-x: An embodied multimodal action model with grounded chain of
  thought and look-ahead spatial reasoning.
\newblock \emph{arXiv preprint arXiv:2412.11974}, 2024.

\bibitem[Tang et~al.(2022)Tang, Liu, Qian, Wu, and Wang]{Tang_2022_CVPR}
Jiaqi Tang, Zhaoyang Liu, Chen Qian, Wayne Wu, and Limin Wang.
\newblock Progressive attention on multi-level dense difference maps for
  generic event boundary detection.
\newblock In \emph{Proceedings of the IEEE/CVF Conference on Computer Vision
  and Pattern Recognition (CVPR)}, pages 3355--3364, 2022.

\bibitem[Team et~al.(2023)Team, Anil, Borgeaud, Alayrac, Yu, Soricut,
  Schalkwyk, Dai, Hauth, Millican, et~al.]{gemini}
Gemini Team, Rohan Anil, Sebastian Borgeaud, Jean-Baptiste Alayrac, Jiahui Yu,
  Radu Soricut, Johan Schalkwyk, Andrew~M Dai, Anja Hauth, Katie Millican,
  et~al.
\newblock Gemini: a family of highly capable multimodal models.
\newblock \emph{arXiv preprint arXiv:2312.11805}, 2023.

\bibitem[Team et~al.(2024)Team, Ghosh, Walke, Pertsch, Black, Mees, Dasari,
  Hejna, Kreiman, Xu, et~al.]{octo}
Octo~Model Team, Dibya Ghosh, Homer Walke, Karl Pertsch, Kevin Black, Oier
  Mees, Sudeep Dasari, Joey Hejna, Tobias Kreiman, Charles Xu, et~al.
\newblock Octo: An open-source generalist robot policy.
\newblock \emph{arXiv preprint arXiv:2405.12213}, 2024.

\bibitem[Van~Vo et~al.(2025)Van~Vo, Nguyen, Nguyen, Nguyen, and Vu]{refinevla}
Tuan Van~Vo, Tan~Quang Nguyen, Khang~Minh Nguyen, Duy Ho~Minh Nguyen, and
  Minh~Nhat Vu.
\newblock Refinevla: Reasoning-aware teacher-guided transfer fine-tuning.
\newblock \emph{arXiv preprint arXiv:2505.19080}, 2025.

\bibitem[Walke et~al.(2023)Walke, Black, Zhao, Vuong, Zheng, Hansen-Estruch,
  He, Myers, Kim, Du, et~al.]{bridgev2}
Homer~Rich Walke, Kevin Black, Tony~Z Zhao, Quan Vuong, Chongyi Zheng, Philippe
  Hansen-Estruch, Andre~Wang He, Vivek Myers, Moo~Jin Kim, Max Du, et~al.
\newblock Bridgedata v2: A dataset for robot learning at scale.
\newblock In \emph{Conference on Robot Learning}, pages 1723--1736. PMLR, 2023.

\bibitem[Wang et~al.(2025{\natexlab{a}})Wang, Ding, Chen, Wu, Wang, Xie, and
  Zhao]{wang-etal-2025-vidorag}
Qiuchen Wang, Ruixue Ding, Zehui Chen, Weiqi Wu, Shihang Wang, Pengjun Xie, and
  Feng Zhao.
\newblock {V}i{D}o{RAG}: Visual document retrieval-augmented generation via
  dynamic iterative reasoning agents.
\newblock In \emph{Proceedings of the 2025 Conference on Empirical Methods in
  Natural Language Processing}, pages 9124--9145, Suzhou, China,
  2025{\natexlab{a}}. Association for Computational Linguistics.

\bibitem[Wang et~al.(2025{\natexlab{b}})Wang, Ding, Zeng, Chen, Chen, Wang,
  Xie, Huang, and Zhao]{wang2025vrag}
Qiuchen Wang, Ruixue Ding, Yu Zeng, Zehui Chen, Lin Chen, Shihang Wang, Pengjun
  Xie, Fei Huang, and Feng Zhao.
\newblock Vrag-rl: Empower vision-perception-based rag for visually rich
  information understanding via iterative reasoning with reinforcement
  learning.
\newblock \emph{arXiv preprint arXiv:2505.22019}, 2025{\natexlab{b}}.

\bibitem[Wang et~al.(2025{\natexlab{c}})Wang, Ding, Li, Cui, Ge, Tong, Song,
  Zhao, Zhao, Hou, et~al.]{wang2025vla}
Yihao Wang, Pengxiang Ding, Lingxiao Li, Can Cui, Zirui Ge, Xinyang Tong,
  Wenxuan Song, Han Zhao, Wei Zhao, Pengxu Hou, et~al.
\newblock Vla-adapter: An effective paradigm for tiny-scale
  vision-language-action model.
\newblock \emph{arXiv preprint arXiv:2509.09372}, 2025{\natexlab{c}}.

\bibitem[Warren~Jr(1990)]{warren1990perception}
William~H Warren~Jr.
\newblock The perception-action coupling.
\newblock In \emph{Sensory-Motor Organizations and Development in Infancy and
  Early Childhood: Proceedings of the NATO Advanced Research Workshop on
  Sensory-Motor Organizations and Development in Infancy and Early Childhood
  Chateu de Rosey, France}, pages 23--37. Springer, 1990.

\bibitem[Xiong et~al.(2025)Xiong, Wang, Guo, Ye, Fan, Gu, Huang, and
  Li]{xiong2025llava}
Tianyi Xiong, Xiyao Wang, Dong Guo, Qinghao Ye, Haoqi Fan, Quanquan Gu, Heng
  Huang, and Chunyuan Li.
\newblock Llava-critic: Learning to evaluate multimodal models.
\newblock In \emph{Proceedings of the Computer Vision and Pattern Recognition
  Conference}, pages 13618--13628, 2025.

\bibitem[Yang et~al.(2025{\natexlab{a}})Yang, Tan, Wu, Zheng, Peng, Liang, Gu,
  Cai, Ye, Jang, et~al.]{magma}
Jianwei Yang, Reuben Tan, Qianhui Wu, Ruijie Zheng, Baolin Peng, Yongyuan
  Liang, Yu Gu, Mu Cai, Seonghyeon Ye, Joel Jang, et~al.
\newblock Magma: A foundation model for multimodal ai agents.
\newblock In \emph{Proceedings of the Computer Vision and Pattern Recognition
  Conference}, pages 14203--14214, 2025{\natexlab{a}}.

\bibitem[Yang et~al.(2025{\natexlab{b}})Yang, Li, Chen, Wang, Tian, Wang, Wang,
  Zhao, Liao, and Pang]{instructvla}
Shuai Yang, Hao Li, Yilun Chen, Bin Wang, Yang Tian, Tai Wang, Hanqing Wang,
  Feng Zhao, Yiyi Liao, and Jiangmiao Pang.
\newblock Instructvla: Vision-language-action instruction tuning from
  understanding to manipulation.
\newblock \emph{arXiv preprint arXiv:2507.17520}, 2025{\natexlab{b}}.

\bibitem[Yu et~al.(2023)Yu, Yang, Li, Wang, Lin, Liu, Wang, and Wang]{MMVET}
Weihao Yu, Zhengyuan Yang, Linjie Li, Jianfeng Wang, Kevin Lin, Zicheng Liu,
  Xinchao Wang, and Lijuan Wang.
\newblock Mm-vet: Evaluating large multimodal models for integrated
  capabilities.
\newblock \emph{arXiv preprint arXiv:2308.02490}, 2023.

\bibitem[Yue et~al.(2024)Yue, Ni, Zhang, Zheng, Liu, Zhang, Stevens, Jiang,
  Ren, Sun, et~al.]{MMMU}
Xiang Yue, Yuansheng Ni, Kai Zhang, Tianyu Zheng, Ruoqi Liu, Ge Zhang, Samuel
  Stevens, Dongfu Jiang, Weiming Ren, Yuxuan Sun, et~al.
\newblock Mmmu: A massive multi-discipline multimodal understanding and
  reasoning benchmark for expert agi.
\newblock In \emph{Proceedings of the IEEE/CVF Conference on Computer Vision
  and Pattern Recognition}, pages 9556--9567, 2024.

\bibitem[Zawalski et~al.(2024)Zawalski, Chen, Pertsch, Mees, Finn, and
  Levine]{ecot}
Michał Zawalski, William Chen, Karl Pertsch, Oier Mees, Chelsea Finn, and
  Sergey Levine.
\newblock Robotic control via embodied chain-of-thought reasoning.
\newblock \emph{arXiv preprint arXiv:2407.08693}, 2024.

\bibitem[Zeng et~al.(2025{\natexlab{a}})Zeng, Huang, Huang, Bao, Qi, Zhao,
  Wang, Chen, Chen, Chen, et~al.]{zeng2025agentic}
Yu Zeng, Wenxuan Huang, Shiting Huang, Xikun Bao, Yukun Qi, Yiming Zhao,
  Qiuchen Wang, Lin Chen, Zehui Chen, Huaian Chen, et~al.
\newblock Agentic jigsaw interaction learning for enhancing visual perception
  and reasoning in vision-language models.
\newblock \emph{arXiv preprint arXiv:2510.01304}, 2025{\natexlab{a}}.

\bibitem[Zeng et~al.(2025{\natexlab{b}})Zeng, Qi, Zhao, Bao, Chen, Chen, Huang,
  Zhao, and Zhao]{zeng-etal-2025-enhancing-large}
Yu Zeng, Yukun Qi, Yiming Zhao, Xikun Bao, Lin Chen, Zehui Chen, Shiting Huang,
  Jie Zhao, and Feng Zhao.
\newblock Enhancing large vision-language models with ultra-detailed image
  caption generation.
\newblock In \emph{Proceedings of the 2025 Conference on Empirical Methods in
  Natural Language Processing}, pages 26703--26729, Suzhou, China,
  2025{\natexlab{b}}. Association for Computational Linguistics.

\bibitem[Zhang et~al.(2024)Zhang, Zhao, Ying, Ma, and Lee]{zhang2024omagen}
Lu Zhang, Tiancheng Zhao, Heting Ying, Yibo Ma, and Kyusong Lee.
\newblock Omagent: A multi-modal agent framework for complex video
  understanding with task divide-and-conquer.
\newblock \emph{arXiv preprint arXiv:2406.16620}, 2024.

\bibitem[Zhang et~al.(2025)Zhang, Zheng, Wu, Zhang, Lin, Yu, Liu, Zhou, and
  Lin]{zhang2025lessons}
Zhenru Zhang, Chujie Zheng, Yangzhen Wu, Beichen Zhang, Runji Lin, Bowen Yu,
  Dayiheng Liu, Jingren Zhou, and Junyang Lin.
\newblock The lessons of developing process reward models in mathematical
  reasoning.
\newblock \emph{arXiv preprint arXiv:2501.07301}, 2025.

\bibitem[Zhao et~al.(2025)Zhao, Zeng, Qi, Liu, Chen, Chen, Bao, Zhao, and
  Zhao]{zhao2025v2p}
Yiming Zhao, Yu Zeng, Yukun Qi, YaoYang Liu, Lin Chen, Zehui Chen, Xikun Bao,
  Jie Zhao, and Feng Zhao.
\newblock V2p-bench: Evaluating video-language understanding with visual
  prompts for better human-model interaction.
\newblock \emph{arXiv preprint arXiv:2503.17736}, 2025.

\bibitem[Zheng et~al.(2023)Zheng, Chiang, Sheng, Zhuang, Wu, Zhuang, Lin, Li,
  Li, Xing, et~al.]{llm-as-a-judge}
Lianmin Zheng, Wei-Lin Chiang, Ying Sheng, Siyuan Zhuang, Zhanghao Wu, Yonghao
  Zhuang, Zi Lin, Zhuohan Li, Dacheng Li, Eric Xing, et~al.
\newblock Judging llm-as-a-judge with mt-bench and chatbot arena.
\newblock \emph{Advances in neural information processing systems},
  36:\penalty0 46595--46623, 2023.

\bibitem[Zheng et~al.(2024)Zheng, Liang, Huang, Gao, Daum{\'e}~III, Kolobov,
  Huang, and Yang]{tracevla}
Ruijie Zheng, Yongyuan Liang, Shuaiyi Huang, Jianfeng Gao, Hal Daum{\'e}~III,
  Andrey Kolobov, Furong Huang, and Jianwei Yang.
\newblock Tracevla: Visual trace prompting enhances spatial-temporal awareness
  for generalist robotic policies.
\newblock \emph{arXiv preprint arXiv:2412.10345}, 2024.

\bibitem[Zhong et~al.(2025)Zhong, Bai, Cai, Huang, Chen, Zhang, Wang, Guo,
  Guan, Lui, et~al.]{zhong2025survey}
Yifan Zhong, Fengshuo Bai, Shaofei Cai, Xuchuan Huang, Zhang Chen, Xiaowei
  Zhang, Yuanfei Wang, Shaoyang Guo, Tianrui Guan, Ka~Nam Lui, et~al.
\newblock A survey on vision-language-action models: An action tokenization
  perspective.
\newblock \emph{arXiv preprint arXiv:2507.01925}, 2025.

\bibitem[Zhou et~al.(2025)Zhou, Zhu, Zhu, Wen, Liu, Xu, Meng, Cheng, Peng,
  Shen, et~al.]{chatvla}
Zhongyi Zhou, Yichen Zhu, Minjie Zhu, Junjie Wen, Ning Liu, Zhiyuan Xu, Weibin
  Meng, Ran Cheng, Yaxin Peng, Chaomin Shen, et~al.
\newblock Chatvla: Unified multimodal understanding and robot control with
  vision-language-action model.
\newblock \emph{arXiv preprint arXiv:2502.14420}, 2025.

\bibitem[Zhuge et~al.(2024)Zhuge, Zhao, Ashley, Wang, Khizbullin, Xiong, Liu,
  Chang, Krishnamoorthi, Tian, et~al.]{zhuge2024agent}
Mingchen Zhuge, Changsheng Zhao, Dylan Ashley, Wenyi Wang, Dmitrii Khizbullin,
  Yunyang Xiong, Zechun Liu, Ernie Chang, Raghuraman Krishnamoorthi, Yuandong
  Tian, et~al.
\newblock Agent-as-a-judge: Evaluate agents with agents.
\newblock \emph{arXiv preprint arXiv:2410.10934}, 2024.

\bibitem[Zitkovich et~al.(2023)Zitkovich, Yu, Xu, Xu, Xiao, Xia, Wu, Wohlhart,
  Welker, Wahid, et~al.]{rt2}
Brianna Zitkovich, Tianhe Yu, Sichun Xu, Peng Xu, Ted Xiao, Fei Xia, Jialin Wu,
  Paul Wohlhart, Stefan Welker, Ayzaan Wahid, et~al.
\newblock Rt-2: Vision-language-action models transfer web knowledge to robotic
  control.
\newblock In \emph{Conference on Robot Learning}, pages 2165--2183. PMLR, 2023.

\end{thebibliography}
